\documentclass[11pt]{article}
\usepackage[margin=1in]{geometry}

\usepackage{amsmath,amssymb}
\usepackage{cite}
\usepackage{hyperref}
\usepackage{graphicx}
\usepackage{longtable,caption}

\newcommand{\Daicc}{\Delta{\rm AIC}_{\rm c}}

\renewcommand{\Re}{\mathrm{Re}}

\begin{document}

\begin{flushleft}
{\LARGE
\textbf{How individuals change language} }

Richard A Blythe\textsuperscript{1*},
William Croft\textsuperscript{2}.

\bigskip
1. SUPA, School of Physics and Astronomy, University of Edinburgh, Edinburgh, UK

2. Department of Linguistics, University of New Mexico, Albuquerque, New Mexico, USA

\bigskip

* r.a.blythe@ed.ac.uk

\end{flushleft}

\section*{Abstract}
Languages emerge and change over time at the population level though interactions between individual speakers. It is, however, hard to directly observe how a single speaker's linguistic innovation precipitates a population-wide change in the language, and many theoretical proposals exist. We introduce a very general mathematical model that encompasses a wide variety of individual-level linguistic behaviours and provides statistical predictions for the population-level changes that result from them. This model allows us to compare the likelihood of empirically-attested changes in definite and indefinite articles in multiple languages under different assumptions on the way in which individuals learn and use language. We find that accounts of language change that appeal primarily to errors in childhood language acquisition are very weakly supported by the historical data, whereas those that allow speakers to change incrementally across the lifespan are more plausible, particularly when combined with social network effects.

\section{Introduction}

Human language is a multiscale phenomenon. A language is shared by a large population, that is, the speech community: it is a set of linguistic conventions, characteristic of the population as a whole. Yet language originates in individuals. Individuals in a population use language to achieve specific communicative goals, and through repeated interactions there emerge the linguistic conventions of the speech community. These conventions also change over time, and as speech communities split, the linguistic conventions of the speech communities diverge, leading to variation across languages. 

How does the behaviour of individual speakers lead to change in linguistic conventions and ultimately the emergence of linguistic diversity?  It transpires that this is one of the most debated questions in the study of language change for at least a century \cite{Jespersen1922}. A widely-held view is that the locus of language change is in child language acquisition, in particular the process of inferring a grammar that is consistent with the sentences that have been heard \cite{Halle1962,Niyogi1997,Yang2000,Lightfoot2013}. Where these sentences do not fully specify a grammar, a child can infer a different grammar from its parents. If enough children infer a different grammar, then the language changes as the generations succeed each other. Variations on this basic idea exist, for example, where a child may have multiple grammars representing old and new linguistic variants, with the relative weighting of the two grammars shifting across generations~\cite{Yang2000}. A competing account is the usage-based theory \cite{Barlow2000,Croft2000,Bybee2010,Bybee2015}, where linguistic innovation occurs at any point in a speaker's lifespan, and speakers vary the frequencies that they use different structures incrementally across the lifespan \cite{Labov2001,Nevalainen2003,Sankoff2007,Baxter2016}.

One reason that this question has not been resolved during the century-long debate is that direct evidence of the origin of a change that develops into a new linguistic convention is generally lacking. Research in child language acquisition has demonstrated that children are very good at acquiring and conforming to the conventions of the speech community. In fact, the primary research question in child language acquisition is how children are so successful in mastering not only general rules of language but also the many exceptions and irregularities in adult language conventions \cite{Bowerman1987}. Child-based approaches argue that children find the patterns rapidly on the basis of specific innate language structures, while usage-based approaches argue that child language acquisition is incremental and general patterns are expanded gradually \cite{Tomasello2003}. The fate of any innovations that are produced in the acquisition phase tends not to be investigated in this line of research. Meanwhile, sociolinguistic research on variation and change begins with a situation in which the novel variant has already been produced, and in fact the novel variant is already changing in frequency on the way to becoming a new linguistic convention. It is virtually impossible to capture the innovation as it happens; linguists are always analysing situations in which the new variant is already present.

Hence linguists have tended to rely on indirect evidence that would shed light on the role of the individual in language change. For example, it has been observed that the sound changes that are produced by children---innovations, or ``errors'' from the perspective of adult grammar---are not the same as the sound changes that have been documented in language history \cite{Dressler1974,Drachman1978,Vihman1980,Hooper1980,Bybee1982,Slobin1997}. However, the innovative variation produced spontaneously by adults in both sound and grammar \emph{is} of the same type that has been documented in language history \cite{Ohala1989,Croft2010}. These observations support the usage-based theory over the child-based theory. Also, while children are extremely good at acquiring the linguistic conventions of adults, by late adolescence they develop into the leaders propagating a novel variant through the speech community, which suggests that language change does not originate in childhood \cite{Labov2001,Tagliamonte2007,Tagliamonte2009,Baxter2016}. 

Here we take a novel approach to addressing the question of the locus of language change in the individual: we quantify and compare the plausibility of different theories of individual behaviour in producing population-level language changes and the resultant worldwide diversity of language traits. We achieve this by introducing a mathematical model that allows us to test a variety of hypotheses about how individuals ultimately bring about language change at the population level. The model is applied to diachronic and crosslinguistic data of one common type of language change, the grammatical evolution of definite and indefinite articles, such as English \emph{the} and \emph{a} respectively. The evolution of articles can be analysed as a cycle of states in which a language without an article may develop an article which may then disappear, allowing a simple unidirectional model of innovation and propagation of a change in a finite set of states. We draw on data of attested changes in definite and indefinite articles for 52 languages, and on the cross-linguistic distribution of article states (620 languages for definite articles, 534 languages for indefinite articles; see below for further details).

Our model allows us to access a very wide range of different individual-level processes of language learning and use which appear in different combinations, whilst remaining amenable to mathematical analysis with methods from population genetics \cite{McCandlish2014}. Specifically, we can estimate the likelihood of our set of empirical language changes at the population scale, given a certain set of assumptions on the behaviour at the individual level. This then means we can determine the regions within this model space that have the strongest empirical support. As we will show below, we find that explanations of language change that appeal exclusively to childhood language learning receive considerably less support than those that allow incremental change across the lifespan. Our analysis further suggests that the complex structure of social networks---in which the degree of influence that different speakers may have over others is highly variable---may play an important role in the diffusion of linguistic innovations.

\section{Data and methods}

\begin{figure}
\centering
\includegraphics[width=.9\linewidth]{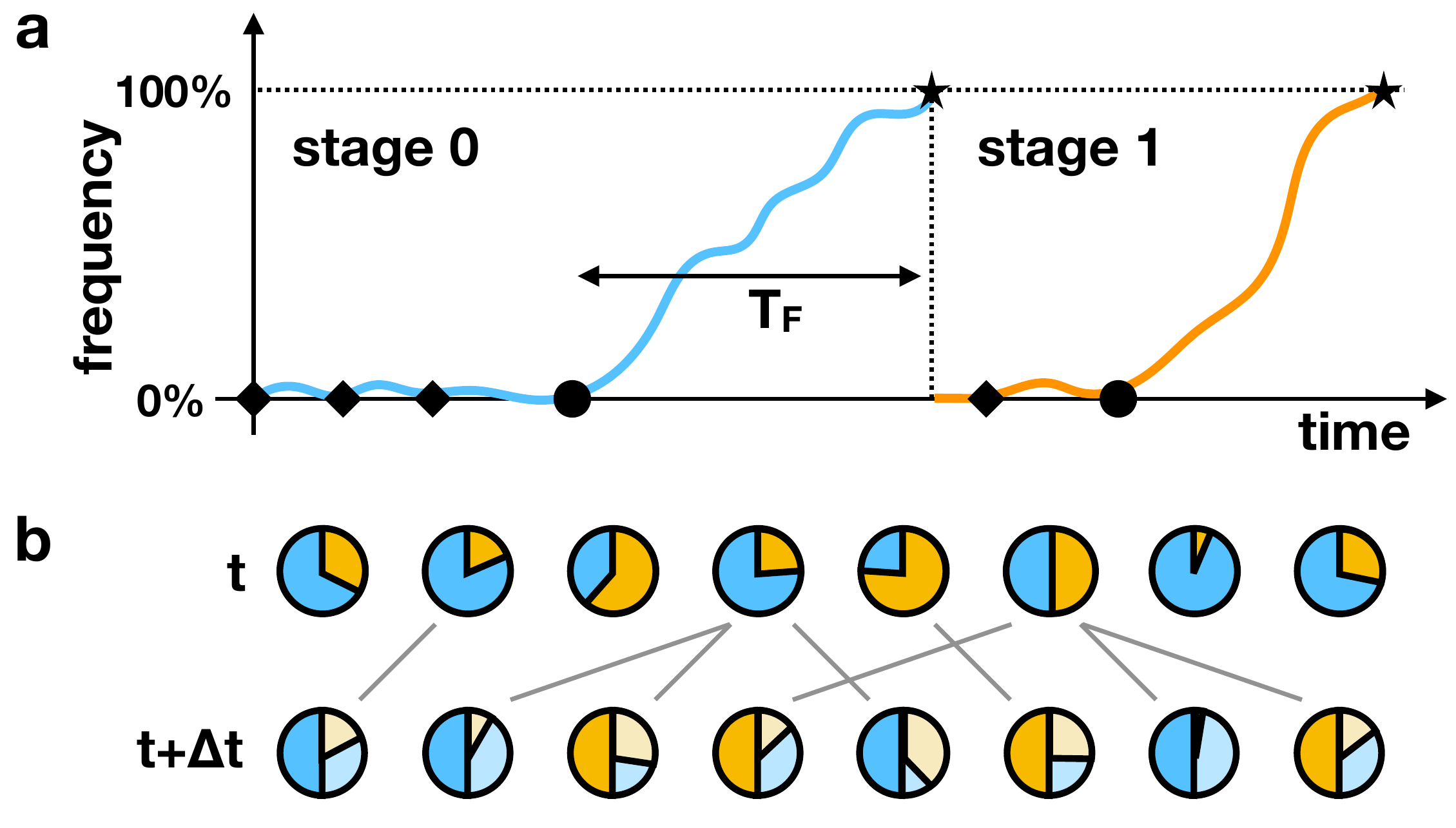} 
\caption{\label{fig:WF-OF} (a) Origin-fixation model at the population scale, showing a transition between two stages of a grammaticalisation cycle (set out in Table~\ref{tab:wals}). Innovations are repeatedly introduced to the population; most fail (diamonds), but some successfully originate a change that propagates and goes to fixation (circles). The fixation time $T_F$ is a random variable (see text). (b) Underlying individual-based (Wright-Fisher) model. Individuals are characterised by the frequency with which they use the innovation (orange portion of pie charts). In the case shown, individuals update their innovation frequencies by retaining a fraction $1-\epsilon$ their existing value, and acquiring the remaining fraction $\epsilon$ through exposure to one other member of the speech community. In the figure, $\epsilon=\frac{1}{2}$ for illustrative purposes. The two levels of description are connected by averaging over the individual speaker-level innovation frequencies in the Wright-Fisher model to obtain the population-level frequency plotted for the origin-fixation model.}
\end{figure}

In this section we first set out empirical properties of changes in articles that guide us towards a statistical model of language change over historical time at the population scale. 
The basic picture, illustrated in Fig~\ref{fig:WF-OF}a, is one in which the population is initially at some stage of the cycle, for example, the situation where there is no definite article (stage 0). As a consequence of individual speaker innovations, an article is occasionally introduced into the population by recruiting a pre-existing word for the article function. This is indicated by diamonds in the figure. In later stages, different linguistic processes lead to a divergence in form, reduction of that form to an affix and the loss of the form. Eventually, one of the innovations propagates so that its \emph{frequency}, defined as the proportion of relevant contexts in which the innovation is used, rises to 100\%. Once this occurs, the next stage of the cycle has been reached and the process begins afresh. Following \cite{McCandlish2014}, we refer to this population-scale model as an \emph{origin-fixation model}: the introduction of an innovation that successfully propagates (denoted by a circle in the figure) is referred to as \emph{origination}, and the point at which it reaches a frequency of 100\% is called \emph{fixation}.

This population-scale process is the product of interactions between individual speakers in the population, that is, acquisition or use, or a combination of the two. These interactions are illustrated schematically in Fig~\ref{fig:WF-OF}b and will be discussed in detail in the second part of this section. The individual-based model is very similar to the \emph{Wright-Fisher model} in population genetics (see e.g.~\cite{Crow1970}), and we refer to it as such. In this model, each speaker is characterised by the frequency with which they use an innovation in the relevant linguistic context. The Wright-Fisher and origin-fixation models are connected by averaging over the individual frequencies to obtain the corresponding frequency at the population level. This then provides a quantitative model for language change over historical timescales that is grounded in individual speaker interactions.

\subsection{Language change at the population level}

\subsubsection{Empirical properties}

We draw on two sources of data to characterise language change at the population level: (i) a survey of documented instances of historical language change (detailed in Appendix~\ref{app:data}); and (ii) the typological distribution of the current stage in the cycle across the world's languages (as recorded in the World Atlas of Language Structures, WALS \cite{WALS2013}). As stated in the Introduction, we focus on definite and indefinite articles for this analysis. There are a number of reasons for this. First, the evolution of articles predominantly follows a single cycle of grammaticalisation. Definite articles are predominantly derived from demonstratives such as \emph{that} \cite{Greenberg1978}, and indefinite articles are predominantly derived from the numeral \emph{one} \cite{Givon1981}. Both articles proceed to being affixed and then disappear. Second, articles are unstable: several find articles to rank among the least stable of a large set of features \cite{Wichmann2009,Dediu2010,Kauhanen2018}. This means that our historical survey includes many documented instances of multiple stages in the article grammaticalisation cycle, which in turn leads to a more sensitive likelihood-based analysis than is possible when changes are rare. Finally, this instability implies that the current distribution of stages in the cycle across languages is likely to be close to the stationary distribution, which simplifies the analysis. Although articles are at one end of the stability spectrum, we expect that similar results to those reported below would be found for more stable features: we return to this point in the Discussion.

We divide the stages of the cycle following the classification of WALS Features 37A and 38A \cite{WALS2013}: (0) no explicit article; (1) use of \emph{that} and \emph{one} for definite and indefinite article meaning respectively; (2) use of a distinct word usually derived from \emph{that} or \emph{one} for the article; and (3) use of an affix. WALS provides the current crosslinguistic distribution of these four stages for definite and indefinite articles (see Table~\ref{tab:wals}). One can also look at the joint distribution of the two features to establish whether they are correlated. A $\chi^2$ test on the contingency table indicates that the features are unlikely to be independent ($p<10^{-6}$; although the conditions for the validity of the $\chi^2$ test do not strictly apply, this level of significance was confirmed by a Monte Carlo sampling procedure).

\begin{table}
\centering
\begin{tabular}{|r|l|r|l|r|}
\hline
& \multicolumn{2}{c|}{Definite}&\multicolumn{2}{c|}{Indefinite}\\
\hline
State & Description & Number & Description & Number \\
\hline
0 & No article & 243 & No article & 296 \\ \hline
1 & Same as \emph{that} & 69 & Same as \emph{one} & 112 \\ \hline
2 & Distinct word & 216 & Distinct word & 102 \\ \hline
3 & Affix & 92 & Affix & 24 \\ \hline
\end{tabular}
\caption{\label{tab:wals} Typological distribution of definite and indefinite articles. The number of languages in each state is taken from \cite{WALS2013}.}
\end{table}

\begin{figure}
\centering
\includegraphics[width=.9\linewidth]{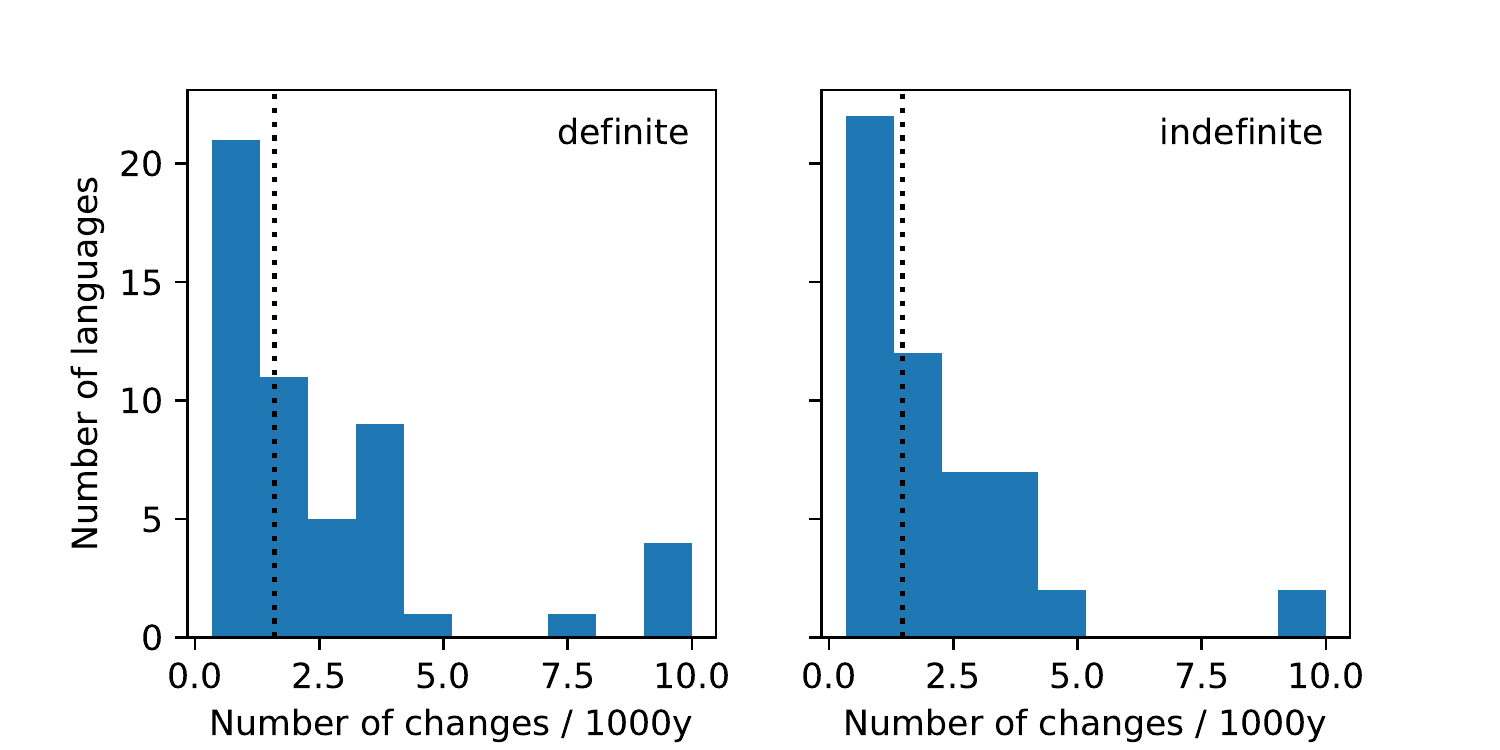} 
\caption{\label{fig:changehist} Distribution of the number of changes in the definite (left) and indefinite (right) article per 1000 years over the empirical dataset of 52 languages. The vertical dotted line indicates the median of the distribution.}
\end{figure}

We collected data on the documented history of articles in 52 languages from multiple sources (see Appendix~\ref{app:data}), and divided their history into the same four stages. Importantly, at any given point in time, one of these conventions typically dominates; over time the dominant convention changes to the next in the sequence 0--3 above, before returning to stage 0 via loss of the article. In our analysis of the 52 languages, we find only a single instance of a stage of the cycle that was skipped. For each article and language, we can estimate the rate of change as $\frac{m+1}{t}$, where $m$ is the number of changes observed and $t$ is the observation period. (Technically, this is the mean of the posterior distribution over rates when the prior is uniform and the changes assumed to occur as a Poisson process). We plot the distribution of these rates for each article in Fig~\ref{fig:changehist}. This shows that the median rate of change is roughly once every 1000 years and that the distribution is somewhat skewed towards slower rates of change. Our survey further suggests that the time taken for a change to propagate is somewhat shorter than this, perhaps of the order of 100 years. We further find that, for any given language, the number of changes in one article is not independent of the other ($\chi^2$ test $p=0.00058$; Monte Carlo estimate $p=0.0026$). In the following we  present results for the two articles separately, as combining probabilities from the two analyses is not justified when measurements are correlated.

\subsubsection{Origin-fixation model}

We use the historical properties of article grammaticalisation cycles, set out above, to flesh out our statistical model of the process at the population scale. Recall from Fig~\ref{fig:WF-OF}a the picture of an initial state in which all speakers are at a given stage of the cycle (say, stage $0$), and as speakers interact, instances of the next stage are repeatedly introduced. In a child-based model \cite{Halle1962,Yang2000,Lightfoot2013}, the next convention is introduced by children in the acquisition process. In the usage-based model, by contrast, the next convention is introduced in language use by speakers of any age \cite{Ohala1989,Croft2000,Bybee2015}. 

Under whatever mechanism one has in mind, only some of the individual innovations are replicated sufficiently often that they become used by the entire population, reaching the frequency of 100\% that defines the state of fixation and therewith the onset of the next stage of the cycle \cite{Ohala1989,Pierrehumbert2003,Croft2010}.

We assume that the rate at which speakers introduce a specific innovation (e.g., introducing a particular form for an article) in individual instances of acquisition or use is constant over time, as is the probability that this innovation then propagates and reaches fixation. This means that at any given stage in the cycle, origination events occur at a constant rate. In mathematical terms, origination is a Poisson process with rate $\omega_i$ when the population is in stage $i$ of the cycle (and so the innovations correspond to stage $i+1$).

Specifically, we take $\omega_i = \frac{\bar{\omega}}{4f_i}$, where $f_i$ is the fraction of languages currently at stage $i$ in the cycle (Table~\ref{tab:wals}). This choice ensures, for any value of the parameter $\bar{\omega}$, that the stationary distribution of the origin-fixation is one in which the probability of being at stage $i$ of the cycle is $f_i$, and consequently matches the WALS distribution (although our conclusions do not depend on this being the case). By including the factor $4$ (i.e., the number of stages in the cycle) $\bar{\omega}$ can be interpreted as a mean origination rate obtained by averaging over one complete cycle. In general we will treat this rate as a free parameter (see Results, below).

Once the originating innovation has entered the population, it takes a time $T_F$, called the \emph{fixation time} to become adopted as the convention by all speakers in the population.
In origin-fixation models applied to the invasion of mutant genes in a biological population \cite{Gillespie1983,McCandlish2014}, the origination process is generally much slower than the fixation process, and $T_F$ is typically set to zero. This is not appropriate in the application to language change: the historical survey above suggests that $T_F$ is only one order of magnitude smaller than the time between the origination of a change. Moreover, $T_F$ is unlikely to be exactly the same for each change, due to the unpredictability of human interactions and individual speech acts.

\begin{figure}
\centering
\includegraphics[width=.9\linewidth]{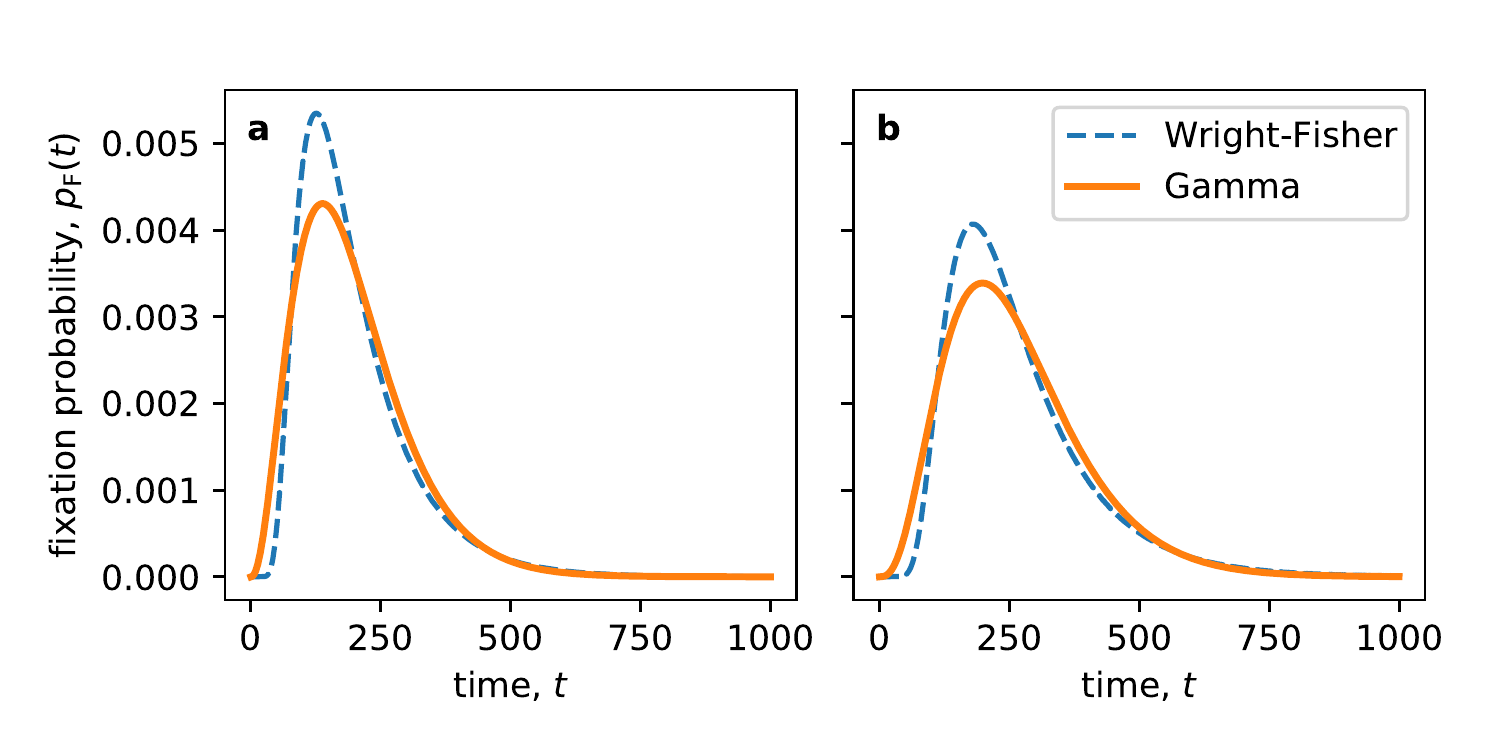} 
\caption{\label{fig:gamma} Approximation of the fixation time distribution obtained numerically for the Wright-Fisher model (dashed line) with a Gamma distribution given by Eq~\ref{gamma}. In (a) the Wright-Fisher model has $N=100$ individuals and no selection. In (b) $N=150$ and $s=0.01$.}
\end{figure}

We account for this unpredictability by drawing each fixation time $T_F$ from a probability distribution. The fixation time distribution can be calculated for certain individual-based models, such as the Wright-Fisher model set out below \cite{Kimura1969,Crow1970}. However, the mathematical form is too complicated to be of practical use, so we approximate it by the simpler Gamma distribution. This distribution is a natural choice for a quantity that is required to be positive (like a fixation time), and whose mean and variance can be controlled independently. In fact, we will arrive at the population-scale model by setting these two quantities equal to those that derive from an underlying individual-based model. Fig~\ref{fig:gamma} shows the Gamma-distribution approximation to the fixation time distribution obtained numerically for the Wright-Fisher model with and without a selection bias. Although the Gamma distribution does not fit perfectly, it captures the location and width of the peak well, and is preferable to simply assuming that $T_F$ is zero.

We now provide a formal mathematical definition of the origin-fixation model that is equivalent to the verbal description above. Starting from stage $i$ of the cycle, a time $T_{O,i}$ at which a change to the next stage in the cycle is originated is drawn from the exponential distribution
\begin{equation}
P_{O,i}(T_{O,i}) = \omega_i {\rm e}^{-\omega_i T_{O,i}}
\end{equation}
as is appropriate for a Poisson process. Then, the time $T_F$ from origination to fixation is drawn from the Gamma distribution
\begin{equation}
\label{gamma}
P_F(T_F) = \frac{\beta^{\alpha}}{\Gamma(\alpha)} T_F^{\alpha-1} {\rm e}^{-\beta T_F}
\quad\mbox{where}\quad \alpha= \frac{\overline{T_F}^2}{\sigma_F^2} \quad\mbox{and}\quad
\beta = \frac{\overline{T_F}}{\sigma_F^2} \;.
\end{equation}
At this point, stage $i+1$ is entered, and origination of a change to stage $i+2$ can begin (by sampling a Poisson process and Gamma-distributed fixation time, as above).

The crucial point is that once these distributions are specified, one can compute the likelihood of the observed changes in our historical survey for any desired combination of parameters $\omega_i$, $\overline{T_F}$ and $\sigma_F^2$. Specifically, we ask for the probability that a language in stage $i$ at the beginning of the observation period reaches stage $j$ by the end of that period. The set of periods, changes, and procedure for calculating the likelihood are detailed in the appendices. In the likelihood calculation, each language is treated as independent of the others: we do however consider a mother and its daughters after a split as separate languages, so that changes in the mother language are not included multiple times in the sample.  It is important to note that the origin-fixation parameters are \emph{not} arbitrary, but depend on the underlying behaviour of individuals. A specific choice of individual-based model will lead to specific values of the parameters $\omega_i$, $\overline{T_F}$ and $\sigma_F^2$, as we establish below.

\subsection{Language change at the individual level}

\subsubsection{Wright-Fisher model}

We now set out a model of language behaviour at the individual level which allows us to determine parameter values for the origin-fixation model in regimes of interest. We start with the fact that all theories of language learning and use involve the linguistic behaviour of one individual in the population being adopted (in some way) by another. Looking backwards in time, one can construct a `genealogy' that shows who acquired linguistic behaviour from whom, parallel to the inheritance of genetic material under biological reproduction. It is well understood in population genetics that many superficially different individual-based models of inheritance generate a common distribution of genealogies \cite{Nordborg2019}. Therefore, one obtains a generic and robust description of an evolutionary process by selecting a specific individual-based model that is adapted to the context at hand. Here we construct a model of the Wright-Fisher type \cite{Crow1970} that allows us to manipulate key properties of the individual speaker, such as how often they can change their behaviour (though learning or use, as appropriate), whether biases towards or against the innovation are operating, and which other members of the speech community they interact with.

The basic structure of this model is shown in Fig~\ref{fig:WF-OF}b. Each circle in the figure represents an individual's linguistic behaviour at a given point in time. Each uses the existing convention (stage 0 in the figure) some fraction of the time, and the incoming innovation (stage 1) the remaining fraction of the time. As in the origin-fixation model, we assume that at most two linguistic variants are widely used at any given time. A variable $x_n$ specifies the relative frequency (in the range $0$ to $1$ inclusive) that speaker $n$ uses the innovation. For example, the left-most speaker in the figure is using the innovation in around $x_1=\frac{1}{3}$ of the relevant contexts at time $t$. In this work, we take $x_n$ to be an average over occurrences of a particular form of the article in a general Noun Phrase construction that expresses (in)definiteness of the referent of the Noun Phrase. The forms are: no article; article identical to a source form (demonstrative for definite article, the numeral `one' for indefinite article); article distinct from source form; and article attached to noun. Although this general construction may be made up of more specific subtypes of Noun Phrase constructions, there is reason to believe that a regular trajectory of change emerges from the aggregation of occurrences over subtypes \cite{Blythe2012}.

 In the traditional Wright-Fisher model, $x_n$ takes only the extremal values $0$ or $1$. In a linguistic context, this corresponds to classic child-based models \cite{Halle1962,Niyogi1997,Lightfoot2013} in which a speaker's grammar is specified in terms of binary parameters. Other models allow for intermediate values of $x_n$: these include variational learning \cite{Yang2000} and usage-based \cite{Tomasello2003} models.

The innovation frequencies $x_n$ are updated at a rate $R$ for each of the $N$ speakers in the population. We define the update rule in a way that includes the child- and usage-based models as special cases. What these have in common is that, in an interaction, each individual is exposed to the behaviour of one other speaker in the population. Each then replaces a fraction $\epsilon$ of their stored linguistic experience with a record of the variant that was perceived in this interaction. That is, $x_n' = (1-\epsilon)x_n + \epsilon \tau$, where $x_n'$ is the updated innovation frequency, and $\tau=1$ if the innovation was perceived in the interaction, and $\tau=0$ otherwise. Fig~\ref{fig:WF-OF}b illustrates this update for the case $\epsilon=\frac{1}{2}$.

The child-based model is obtained when $\epsilon=1$. The update then corresponds to a child being exposed to the behaviour of a parent, applying some learning rule to determine if the grammar of the language corresponds to the convention or the innovation, and setting $x=0$ or $1$ accordingly. Importantly, the learning rule can allow the child to infer a grammar that is different from that of the parent: cue-based learning \cite{Niyogi2009} is one mechanism that allows for this. A general model for such mechanisms can be obtained by introducing a probability $\eta_i$ that, given a behaviour that is consistent with the parent holding grammar $i$ in the cycle, the child nevertheless adopts grammar $i+1$ (for example, because the sentences produced by the parent are more consistent with the next stage of the grammaticalisation cycle). In the child-based model, the appropriate choice for the update rate $R$ would be once per generation. Under these conditions, the timescale of the cultural evolutionary process of language change is necessarily tied to that of biological evolution (although the two processes differ in other respects, for example, the number and identity of parents).

By contrast, the usage-based model allows for the cultural evolutionary dynamics to proceed more quickly than their biological counterparts, as individuals interact  many times in the course of a generation. However, the impact of each interaction is likely to be smaller, implying that the parameter $\epsilon$ that quantifies this impact should be small. Fig~\ref{fig:WF-OF}b illustrates the case of $\epsilon=\frac{1}{2}$, in which after the update (time $t+\Delta t$), half of the usage frequency derives from their behaviour before the interaction (light shading in the figure), and the other half (dark shading) corresponds to whether a conventional or innovative utterance was perceived in an interaction with the speaker shown by the connecting line. As in the child-based model, there is a small probability $\eta_i$ that a conventional behaviour is perceived as an innovation. This can represent a variety of processes that might apply in single instances of use, e.g., auditory and articulatory constraints \cite{Lindblom1983,Christiansen2016} or cognitive biases \cite{StClair2009,Culbertson2012,Christiansen2016}, along with indeterminacy in inferring a phonological form \cite{Ohala1989,Pierrehumbert2003} or meaning \cite{Quine1960,Croft2010}, that may favour one construction over another (see e.g.~\cite{Croft2000} for an extended discussion of innovation in language change).

To complete the description of the Wright-Fisher model, we need to specify how the \emph{interlocutor}---the speaker who provides the linguistic data to the learner (or listener)---is chosen. There are two components to this: (i) a social network structure; and (ii) a possible biasing of interlocutors based on their linguistic behaviour. We describe these in turn.

\begin{figure}
\centering
\includegraphics[width=.9\linewidth]{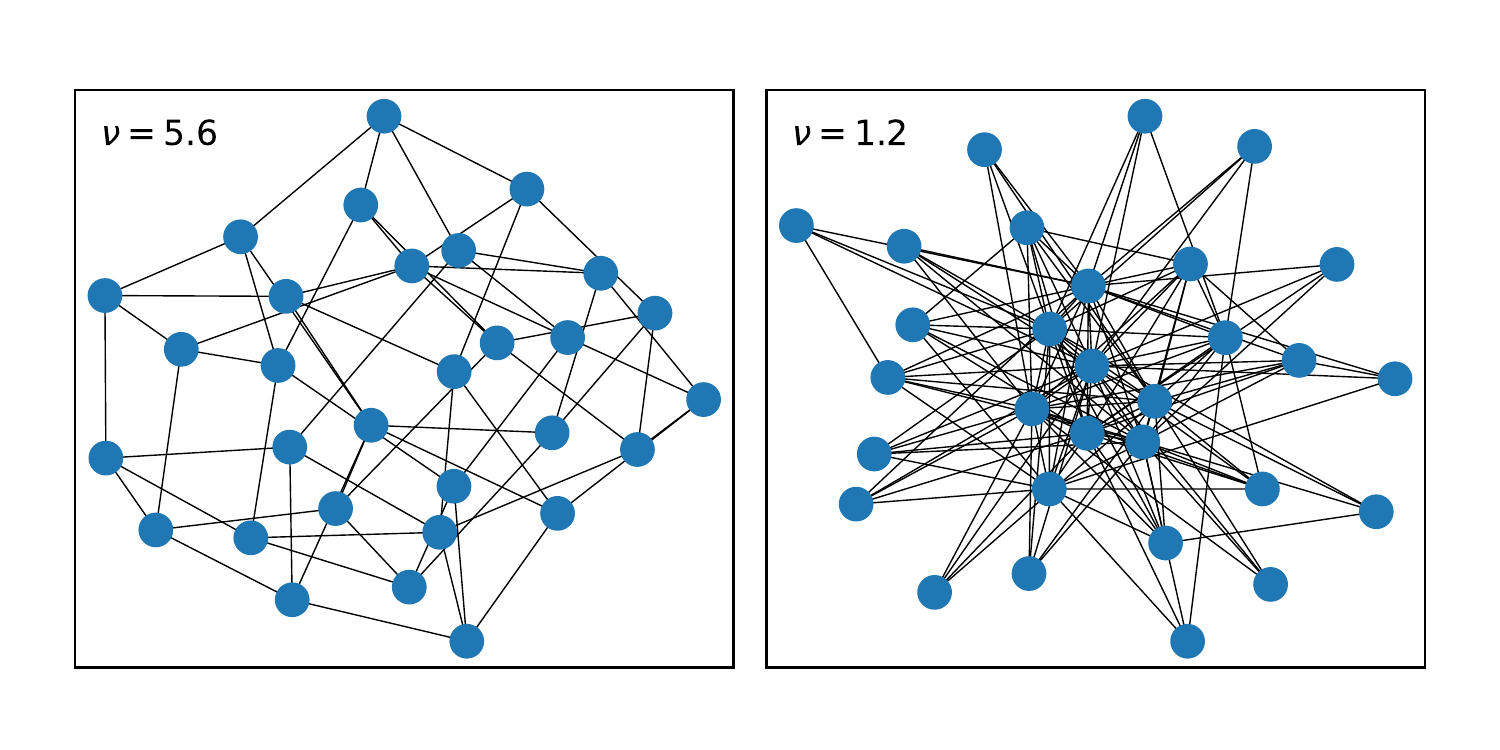} 
\caption{\label{fig:degreedist} Instances of random networks with different degree exponents $\nu$. The case $\nu>2$ (left) corresponds to a homogeneous network in which individuals all have a similar number of neighbours. The case $\nu<2$ (right) is heterogeneous: the central individuals are well-connected whilst the peripheral individuals are not.}
\end{figure}

The social network is set up so that speaker $i$ has $z_i$ immediate neighbours, with $z_i$ drawn from a \emph{degree distribution} $p_z$. Thus different individuals can have different numbers of neighbours. In the absence of the bias, each neighbour is chosen as an interlocutor with equal probability in an interaction. A generic model for social networks is the power-law degree distribution $p_z \propto z^{-(1+\nu)}$ in which the exponent $\nu$ controls the heterogeneity of the network. Values of $\nu>2$ are regarded as homogeneous, in the sense that innovations spread in the population in the same way as on a network in which all speakers have the same number of neighbours (even though there is variation). When $\nu<2$, the networks become increasingly heterogeneous as $\nu$ is decreased: these feature a small number of highly-connected individuals and a large number of relatively isolated individuals. Evolutionary dynamics tend to run faster on heterogeneous networks \cite{Sood2005,Antal2006,Baxter2008}, and there is some evidence that human social networks are heterogeneous ($1.1<\nu<1.3$, \cite{Albert2002,Clauset2009,Kwak2010}). Fig~\ref{fig:degreedist} illustrates the distinction between homogeneous and heterogeneous random networks.

The interlocutor bias is implemented by choosing a neighbour $m$ with a probability proportional to $1 + s x_m$ instead of uniformly. The \emph{selection strength} $s$ serves to favour (if $s>0$) or disfavour (if $s<0$) the innovation, which may originate in one of a number of processes. For example, in the variational learning framework \cite{Yang2000}, there is a systematic bias towards a grammar that parses a larger number of sentences. In a sociolinguistic setting, association between a linguistic variant and a socially prestigious group may lead to a bias towards (or against) that variant \cite{LePage1985,Labov2001}. The case $s=0$ describes a neutral model for language change, which has been discussed in the context of new-dialect formation \cite{Trudgill2000,Baxter2009}.

We emphasise that a large number of models for language learning and use that have been discussed in the literature fall into the Wright-Fisher class, even though they may differ in detail and may not be presented as such. A non-exhaustive list includes those that appeal to cue-based learning \cite{Niyogi2009}, Bayesian learning from one or more teachers \cite{Smith2009,Reali2010,Burkett2010}, variational learning \cite{Yang2000} and usage-based models \cite{Baxter2006}. Moreover, the Wright-Fisher model has been used as a phenomenological model for changes in word frequencies \cite{Newberry2017,Karjus2020,Karsdorp2020}.

We conclude this section with a formal mathematical specification of the Wright-Fisher model. The distribution $P(x,t)$ of the innovation frequency, $x$, at the population level, at a time $t$ after it is originated, is generally well-described by the forward Kolmogorov equation
\begin{equation}
\label{fpe}
T_M \dot{P}(x,t) = -s[x(1-x)P(x,t)]' + \frac{1}{2N_e} [ x(1-x) P(x,t) ]''
\end{equation}
in which a dot and prime denote derivatives with respect to $t$ and $x$, respectively \cite{Crow1970,Blythe2007}. The parameters $T_M$, $s$ and $N_e$ correspond to a memory lifetime, an innovation bias and an effective population size, respectively. We emphasise that this equation applies between successive origination events, and describes the process by which the innovation propagates (rises to $x=1$) or fails (falls to $x=0$). Therefore the origination rate does not appear in this equation. However, it does enter into a correction factor, set out in Appendix~\ref{app:interfere}, that accounts for the possibility that a second origination occurs before either of these endpoints is reached.

The main difference between models within the Wright-Fisher class is how $T_M$, $s$ and $N_e$ relate to the parameters that apply to a specific model. In the present case, which has the set of parameters specified in Table~\ref{tab:parms}, we have $T_M = 1/(R\epsilon)$, $s$ is as specified above and $N_e=N(\overline{z}^2/\overline{z^2}) / \epsilon$ in which $z$ is the number of neighbours a speaker has on the social network, and the overline denotes an average over speakers \cite{Sood2005,Antal2006,Baxter2008}.

\begin{table}
\centering
\begin{tabular}{|l|l|l|l|}
\hline
\multicolumn{2}{|c|}{Wright-Fisher model}&\multicolumn{2}{c|}{Origin-Fixation model}\\ \hline
Symbol & Meaning & Symbol & Meaning \\ 
\hline
$N$ & speech community size & $\omega_i$ & origination rate \\ \hline
$\nu$ & social network heterogeneity & $\overline{T_F}$ & mean fixation time \\ \hline
$R$ & interaction rate  & $\sigma_{F}^2$ & variance in fixation time \\ \hline
$\epsilon$ & interaction impact  & & \\ \hline
$\eta_i$ & innovation rate  & & \\ \hline
$s$ & selection strength  & & \\ \hline
\end{tabular}
\caption{\label{tab:parms} Parameters in the individual-based Wright-Fisher and population-level Origin-Fixation models. The parameters in the Origin-Fixation model that characterise the dynamics at the population scale can all be expressed in terms of those relating to the behaviour of individuals (see Data and Methods).}
\end{table}

In Appendix~\ref{app:demo} we demonstrate that Eq~\ref{fpe} applies more generally than to the specific agent-based model set out here, and furthermore that the quantities $T_M$, $s$ and $N_e$ have a similar interpretation. This is achieved by considering a model that has many additional features---for example, ongoing birth and death of speakers, changes in social network structure and variation in interaction rates between speakers and over time---and showing that the changes in the innovation frequency $x$ over short time intervals are the same as those described by Eq~\ref{fpe}. Therefore the results we present below do not rely on this model being an accurate representation of language learning and use.

\subsubsection{Connection to origin-fixation model}

We connect the individual to the population scale by determining how the parameters in the origin-fixation model (also specified in Table~\ref{tab:parms}) relate to those in the Wright-Fisher model. The origination rates $\omega_i$ are given by the formula
$\omega_i = N R \eta_i Q(\epsilon/N)$, where $N$ is the number of speakers in the speech community, $\eta_i$ is the individual innovation rate per interaction, $R$ is the interaction rate and $Q(x_0)$ is the probability that an innovation goes to fixation starting from some frequency $x_0$. In the Wright-Fisher model, this initial frequency is $x_0=\epsilon/N$, because exactly one speaker uses the innovation with probability $\epsilon$. We then have
\begin{equation}
\label{q}
Q\left(\frac{\epsilon}{N}\right) = \frac{1 - {\rm e}^{-2N_e s \epsilon/N}}{1 - {\rm e}^{-2N_e s}} \;.
\end{equation}
This result is obtained by solving the backward equation that corresponds to Eq~\ref{fpe} (see \cite{Kimura1969,Crow1970} and Appendix~\ref{app:wfmodel}).  We see that the effective population size, $N_e$ (which depends on the actual population size $N$, the update fraction $\epsilon$ and the social network structure) plays an important part in determining the probability that an innovation propagates. It also determines how quickly an innovation may reach fixation. Numerical methods, described in Appendix~\ref{app:ofmodel} with the code available at \cite{code}, are used to determine exactly how the mean and the variance in the fixation time, $\overline{T_F}$ and $\sigma_F^2$, in the origin-fixation model depend on the Wright-Fisher model parameters. Here we note that the characteristic timescale is of order $T_M N_e$ when the bias $s$ is small, and of order $T_M \ln(N_e)$ when it is large, which turns out to have important consequences for the plausibility of the historical data for specific models of language learning and use in our analysis below.

In summary, then, our basic approach is to use the origin-fixation model to determine the likelihood of an observed set of historical language changes. The parameters in this model are obtained from an underlying Wright-Fisher model, so that we may understand---for example---which learning rates, biases and social network structure are more or less well supported by the historical data. As we have argued, our findings do not depend on the detailed structure of the Wright-Fisher model. The crucial component is that a speaker's behaviour can be represented by an innovation frequency $x$, and that this is affected by learning from or using language with other members of the speech community over time.

\section{Results}

We now compare the likelihood of the empirically attested set of language changes (detailed in Appendix~\ref{app:data}) under different assumptions on the underlying behaviour of individuals in the respective populations. An appropriate measure for likelihood comparison is the Akaike Information Criterion, corrected for small sample sizes (${\rm AIC}_c$, \cite{Burnham1998}), as the models we consider have different structures. It is defined as
\begin{equation}
{\rm AIC}_c = 2k - 2\ln({\cal L}) + \frac{2k(k+1)}{n-k-1}
\end{equation}
where $k$ is the number of free parameters in the model, $n$ is the number of observations and ${\cal L}$ is the likelihood of those $n$ observations, as determined from the origin-fixation model. An observation is the sequence of transitions between different stages of a grammaticalisation cycle over a specified historical time period for a given language, as tabulated in Appendix~\ref{app:data}. The number of observations is therefore the number of languages in the sample (52 for both articles).

The difference in the ${\rm AIC}_c$ 
value between two models, denoted $\Daicc$, gives a measure of how much the model with the lower ${\rm AIC}_c$ score is preferred over the other. Models with more free parameters (higher $k$) can be dispreferred even when the data likelihood increases as a result of increasing parameters. For nested models, this increase is inevitable, but for models with different structures, ${\rm AIC}_c$ remains valid as it is based on general information theoretic principles \cite{Burnham1998}. Given two candidate models and a sufficiently large number of observations, ${\rm e}^{\Daicc/2}$ provides an estimate of the probability that the model with the higher ${\rm AIC}_c$ better describes the data than that with the lower value. There is some freedom to choose the value of $\Daicc$ at which one discards the inferior model. In this work we take a value of around $10$ (corresponding to a likelihood ratio of around $150$) as indicative of the model with the higher ${\rm AIC}_c$ becoming too implausible to consider further. However since there is some flexibility in this regard, we will generally show the dependence of $\Daicc$ on model parameters, so one can gauge the scale of the likelihood differences between models.  It is important to note that such model comparisons do not in themselves validate the superior model: for this one needs to consider goodness-of-fit measures as well \cite{Burnham1998}.

We begin by establishing a baseline against which different individual-level mechanisms of language change will be compared. In this baseline model, language changes occur at the population level as a Poisson process. We emphasise from the outset that this is not an individual-based model of language change: changes in the population occur autonomously without reference to individual speakers. Nevertheless this model helps to illustrate our statistical approach and, as we discuss below, it also provides valuable insights into \emph{why} particular individual-based mechanisms are found to provide more or less plausible explanations of historical language changes at the population level.

\subsection{Poisson baseline}

In the baseline model, we assume that a change from stage $i$ to stage $i+1$ of the cycle occurs as a Poisson process at a constant rate $\omega_i = \bar{\omega}/(4f_i)$ in each population, where $f_i$ is the fraction of the world's languages that is currently at stage $i$ of the cycle (Table~\ref{tab:wals}). This factor of $f_i$ ensures that the stationary distribution in the baseline model matches the contemporary WALS distribution. This model is equivalent to the origin-fixation model of Fig~\ref{fig:WF-OF}a, with instantaneous fixation ($T_F=0$). This model has one free parameter, the mean rate of language change, $\bar{\omega}$, which is estimated by maximising the likelihood of the data.

The maximum likelihood value of $\bar{\omega}$, the corresponding ${\rm AIC}_c$, a classical $p$-value and two goodness-of-fit statistics are  presented in Table~\ref{tab:poisson}.  The $p$-value is the probability, within the model, of all possible transitions between stages of the relevant grammaticalisation cycle over the relevant historical period for each language whose likelihood is lower than the transitions that actually occurred. This $p$-value can be interpreted in the usual way, with a low $p$-value indicating a likely departure from the model assumptions.

By itself, an ${\rm AIC}_c$ score (or differences between them) does not furnish any information about how well a particular model fits the data. To gain an insight into goodness-of-fit, we consider the \emph{overdispersion} of two random variables $X$ (specified below) which quantifies the extent to which observed deviations of $X$ from their mean values $\bar{X}$ within the model are consistent with the expected deviations. For a given observation, the overdispersion is defined as $O_X = (X-\bar{X})^2/{\rm Var(X)}$, that is, the ratio of the observed square deviation to its expected value. If the overdispersion is close to $1$, the deviations are as expected, and we conclude that the distribution of $X$ is well-predicted \cite{Burnham1998}. For a given language, the two quantities $X$ are: (i) the total number of language changes in the historical period; and (ii) a binary variable that equals $1$ if at least one change occurred, or $0$ otherwise. We average over all languages in the sample to obtain the single measure that is presented in Table~\ref{tab:poisson}.

\begin{table}
\centering
\begin{tabular}{|r|r|r|}
\hline
& Definite & Indefinite \\
\hline
$\bar{\omega}$ ($\times 10^{-4}\, {\rm yr}^{-1}$) & $6.05$ & $5.67$ \\ \hline
${\rm AIC}_c$ & $128$ & $93.6$ \\ \hline
$p$ & $0.0097$ & $0.16$ \\ \hline
Overdispersion (number of changes) & $2.7$ & $1.1$ \\ \hline
Overdispersion (at least one change)& $1.1$ & $1.0$ \\ \hline
\end{tabular}
\caption{\label{tab:poisson} Fit of a Poisson process to article grammaticalisation histories.
$\bar{\omega}$ is the maximum likelihood rate of change and ${\rm AIC}_c$ the corrected Akaike information criterion. $p$ is the cumulative probability of events less likely than the observation. Overdispersion measures goodness of fit, with values closer to 1 indicating a better fit. $p$ and overdispersion are estimated from $10^6$ Monte Carlo simulations of the process.}
\end{table}

The low overdispersion scores suggest that this baseline model provides a good description of changes in the indefinite article, whilst it performs less well for the definite article. A likely source of this difference is the larger number of languages whose definite article changes rapidly compared to the indefinite article, as can be seen from Fig~\ref{fig:changehist}. It is further possible that assumptions made about the data (for example, that the distribution of articles is stationary, that changes in different languages are independent, or, indeed, that the fixation time can be idealised to zero) do not strictly hold. We also remark that the second overdispersion measure is less sensitive than the first: however, it turns out that this is easier to calculate for individual-based models, and we will take a large deviation of this measure from 1 as providing a strong indication of a poor fit to the data.

It is remarkable that this simple model seems to provide a reasonably good fit to the data, particularly in view of an ongoing discussion about the role of population size in language structure and change \cite{Wichmann2008,Lupyan2010,Nettle2012,Bromham2015} (a point we return to in the Discussion). The Poisson model explicitly assumes that the phenomenological rate of change $\bar{\omega}$ is constant across all populations, and that each language change is able to propagate rapidly from origination to fixation. These observations suggest that we should expect to find more plausible accounts of historical language change in individual-based models whose emergent population-level dynamics share these properties.

\subsection{Child-based models of language change}

We now examine the constraints on the population-level dynamics of language change that arise from assuming that language change occurs primarily through the process of childhood language acquisition (e.g.,~\cite{Halle1962,Yang2000,Briscoe2000,Niyogi2009,Smith2009,Burkett2010,Smith2016}). As noted above, such theories imply that the rate, $R$, at which a grammar can be updated is once per human generation, which we take to be once every 25 years (i.e, $R=0.04{\rm yr}^{-1}$). In the case where learning causes children converge on a single grammar (i.e., categorical use of one of the four article variants), we take $\epsilon=1$. In the case of variational learners (e.g.~\cite{Yang2000}), speakers can entertain mixtures of grammars: this can be realised with $\epsilon<1$. We consider the categorical case first.

The literature on child-based theories rarely refers to population structure. We therefore begin by assuming that populations are homogeneous: that is, that each child learns from roughly the same number of (cultural) parents, and conversely, that each adult provides linguistic input to roughly the same number of (cultural) offspring. Under these conditions, the emergent origination rates and fixation times in each population depends on a core size that is equal to the population's actual size (see Methods). It is therefore necessary for us to estimate the population (speech community) size for each language over the historical period for which empirical data exist. In Appendices~\ref{app:popsize} and \ref{app:langsize}, we set out the procedure that we use to estimate the mean population size for each language over its recorded period of change. This is then used as the core population size for that language in our analysis.

\begin{figure}
\centering
\includegraphics[width=.9\linewidth]{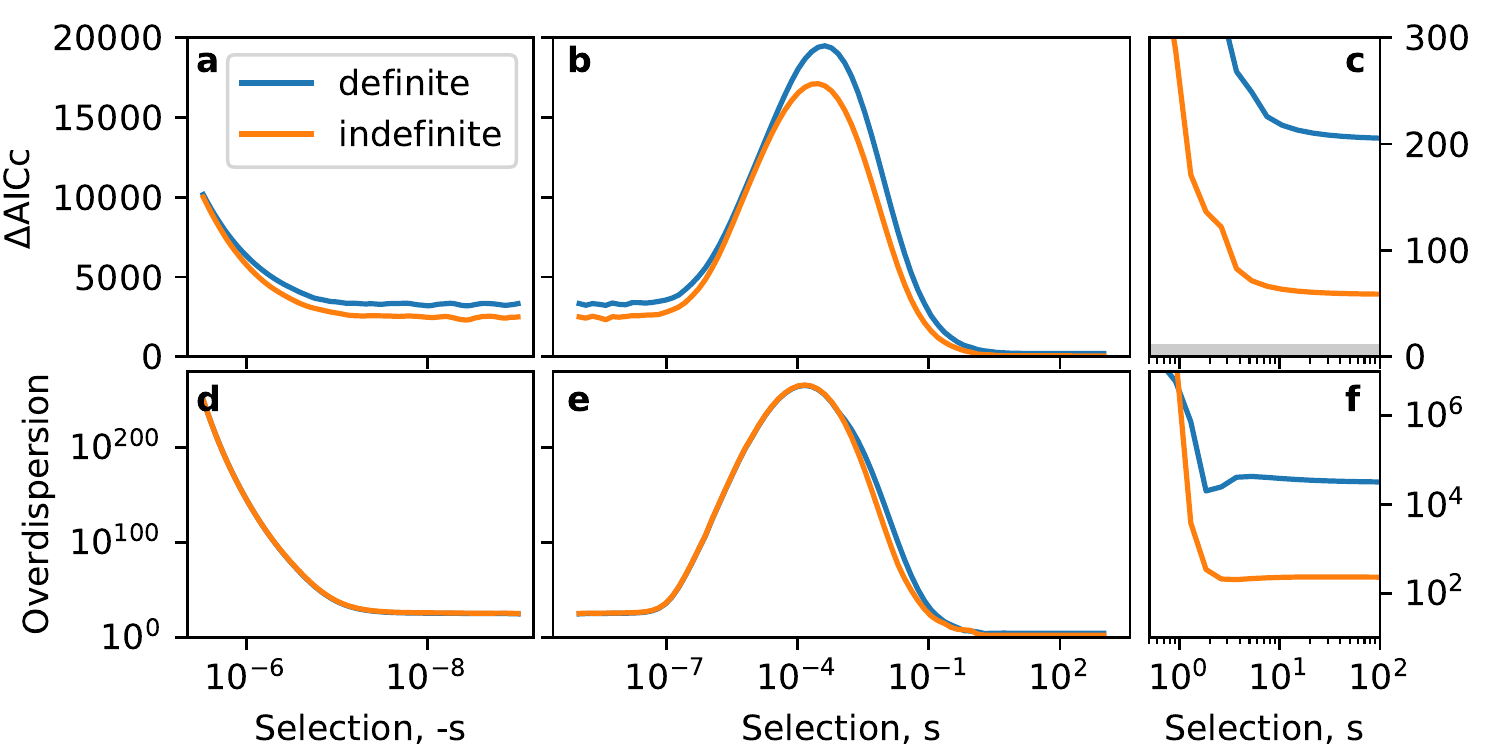} 
\caption{\label{fig:oneshot} $\Daicc$ (panels a--c) and binary overdispersion (c--f) for negative (a and d) and positive (b, c, d and f) selection strength $s$ within a child-based learning paradigm. The smallest values of both measures (which indicate better fits to the data) are obtained for strong positive selection ($s>1$, highlighted in panels c and f which has a larger vertical scale). The $\Daicc$ values are far away from the shaded zone where $\Daicc \le 10$ and the evidence in favour of the child-based model starts to become comparable with that of the baseline.}
\end{figure}

This leaves just two unconstrained parameters, the mean rate $\bar{\eta}$ at which innovations arise in individual instances of language learning (the ``error'' rate, in the child-based model), and the selective bias $s$ in favour of the innovation.  Our strategy is to choose the value of $\bar{\eta}$ that maximises the likelihood of the data set given all other parameter settings, and to plot $\Daicc$ with respect to the Poisson baseline model as a function of the selection strength $s$ so that we can see where the support for the child-based model is strongest. Here, we treat the individual-based model as the candidate model, so $\Daicc = {\rm AIC}_c({\rm candidate}) - {\rm AIC}_c({\rm baseline})$ is positive when the evidence supports the baseline model, and negative when the evidence supports the candidate model. The resulting plot is shown in Fig~\ref{fig:oneshot}, along with a corresponding plot of the second of the two overdispersion measures considered for the Poisson baseline model.

We find that across the entire range of selection strengths $s$, support for the child-based model is very poor. The greatest plausibility (relative to the Poisson baseline) is obtained where $\Daicc$ is smallest: this happens in the limit of infinite selection strength. As can be seen from the rightmost panels of Figure~\ref{fig:oneshot}, the values of $\Daicc$ in these regions are still rather large, reaching asymyptotes at $204$ and $58.4$ for definite and indefinite articles, respectively (both to 3 s.f.). This corresponds to the evidence in favour of the candidate model being $10^{44}$ (definite) and $10^{13}$ (indefinite) times smaller than the baseline.

However this comparison with the Poisson baseline is not entirely fair, as this phenomenological population-level dynamics may not be accessible for any combination of parameters in the individual-based model. For this reason we must also check the goodness-of-fit via the overdispersion measure. Again we find anomalously large values, the asymptotic values being $31300$ (definite) and $226$ (indefinite), suggesting that the assumptions made about the underlying dynamics of language change are wildly inconsistent with the historical data. Throughout this investigation, we found that $\Daicc$ correlates strongly with goodness-of-fit, and so in the rest of this work we show only $\Daicc$, and investigate whether alternative assumptions on the individual-level behaviour are capable of delivering a much smaller $\Daicc$.

To focus this investigation, it is instructive to understand why the empirical data have such a low likelihood (and therewith high $\Daicc$) within the child-based model. As previously noted, the effective population size (which here, is the same as the actual population size) is of fundamental importance in population genetics models \cite{Crow1970}. When the selection strength, $s$, is large, each individual innovation is likely to propagate, and the mean origination rate (at the population level) increases linearly with the population size. On the other hand, when the selection strength is small, the origination rate is roughly constant but fixation time $T_F$ is proportional to the population size.  Since the historical average population sizes in the empirical data set range across six orders of magnitude, then either the origination rate or the fixation time exhibits this wide variation in the child-based model. The fact that the Poisson baseline, which has no dependence on population size at all, apparently provides a much better fit, suggests that individual-based models in which origination rates and fixation times vary more weakly with population size than in the child-based model should be more favoured.  Variants of the child-based model in which grammars are probabilistic \cite{Yang2000} do not fall into this class: these have $\epsilon<1$, which implies a fixation time $N/\epsilon^2$ when $s$ is small. That is, these models are more sensitive to population size than models that allow children to acquire only a single grammar.

\subsection{Usage-based models of language change}

In a usage-based model, a speaker's grammar may change across their lifespan \cite{Tomasello2003}, in principle in response to every utterance they hear (i.e., up to around $10^7$ times a year \cite{Hart1995}).  This has the potential to weaken the sensitivity to population size: if a large number of interactions between speakers  is required for a change to propagate through the population, then the higher interaction frequency in the usage-based model gives the change a greater chance of going through on the attested historical timescales.  However, this effect may be tempered by the fact that the change to each grammar is smaller in each interaction, which has the opposite effect.

\begin{figure}
\centering
\includegraphics[width=.9\linewidth]{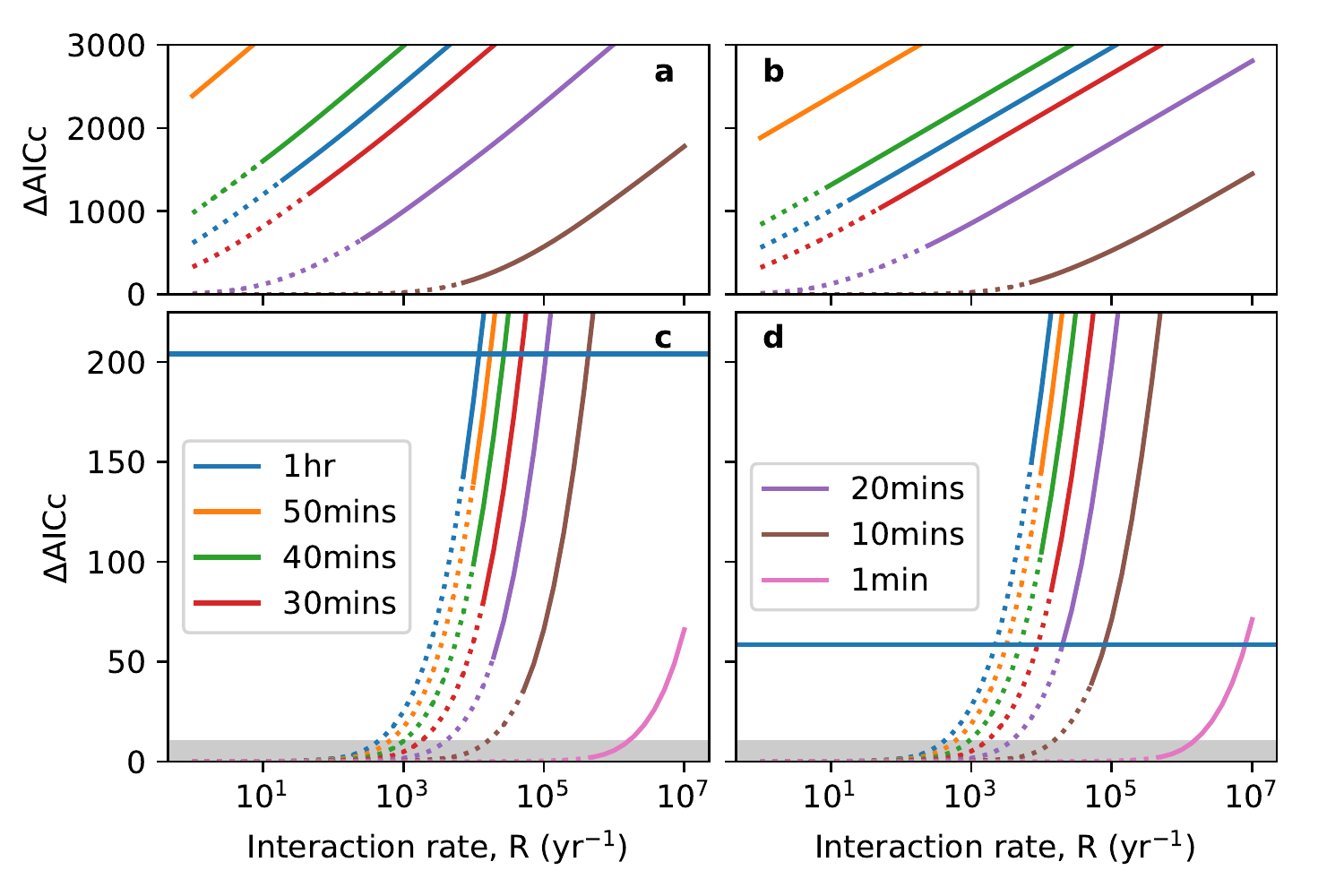} 
\caption{$\Daicc$ in the usage-based model as a function of interaction rate $R$ for the definite (panels a and c) and indefinite (b and d) articles. Along each curve, the memory time $T_M=1/(R\epsilon)$ is held constant. In panels a and b, $T_M$ ranges from $25$ years (top line) to $1$ hour (bottom line). Panels c and d focus on the range of interest where greater plausibility than the child-based model is achieved: the horizontal lines correspond to the $s=\infty$ asymptotes in Fig.~\ref{fig:oneshot}. Dotted lines indicate where the usage-based model is unphysical ($\epsilon>1$) and the shaded grey region indicates where the fit starts to become comparable to the Poisson baseline ($\Daicc<10$).}
\label{fig:usage}
\end{figure}

To explore the interaction between an increased interaction rate $R$, and lower impact on the grammar $\epsilon$, it is convenient to work with the memory time $T_M=1/(R\epsilon)$, which is the expected lifetime of a single item of linguistic experience in the speaker's mind.  Considering again the case of homogeneous populations, we compare in Fig~\ref{fig:usage} the class of usage-based models with no selection ($s=0$) over the reasonable range of $R$ at fixed memory times $T_M=1/(R\epsilon)$ against the baseline model. Note that the dotted parts of the curves correspond to an unphysical parameter value of $\epsilon>1$.  From these $\Daicc$ plots, we see that our intuition that an increased interaction rate allows changes to go through more easily is correct. We achieve greater plausibility than the most plausible child-based model when memory times are short, specifically less than one hour. We note that we can approach the plausibility of the Poisson baseline if we allow $T_M$ to be as short as one minute.

Although shorter memory times in the individual allow for a faster rate of change in the population, the basic property of fixation times being proportional to the population size is unaffected. This is why we find that individual memory times must be very short (perhaps unreasonably so, see Discussion) to improve on child-based models. Furthermore, there is stronger sensitivity to population size when selection is operating ($s\ne0$), which leads to lower plausibility gains with respect to the child-based model than in the neutral case ($s=0$). This suggests that one needs to appeal beyond merely shorter memory times to explain the apparently weak effect of population size on article grammaticalisation cycles.

\subsection{Social network effects}

Studies of the Wright-Fisher and related models on heterogeneous networks \cite{Sood2005,Antal2006,Baxter2008} show that these can weaken the effect of population size on characteristic timescales of change. As discussed in the Wright-Fisher model section, above, we model social networks as those with a power law distribution $P(z) \sim z^{-(1+\nu)}$. We recall that the exponent $\nu$ controls the heterogeneity of the network, with lower values of $\nu$ corresponding to greater heterogeneity: see also Fig~\ref{fig:degreedist}. On such networks, the mean fixation time is proportional to an \emph{effective} population size $N_e \sim N^{2-2/\nu}$ which is less than the actual size $N$ if $1 < \nu < 2$ \cite{Sood2005,Antal2006,Baxter2008}. In the context of language change, we can think of $N_e$ as measuring the size of a core population who exert much greater influence over the periphery than vice versa. Empirical studies of large networks (like friendship networks) provide some support for this power-law distribution with an exponent $\nu$ in the range $1.1<\nu<1.3$ \cite{Albert2002,Clauset2009,Kwak2010}. 

\begin{figure}
\centering
\includegraphics[width=.9\linewidth]{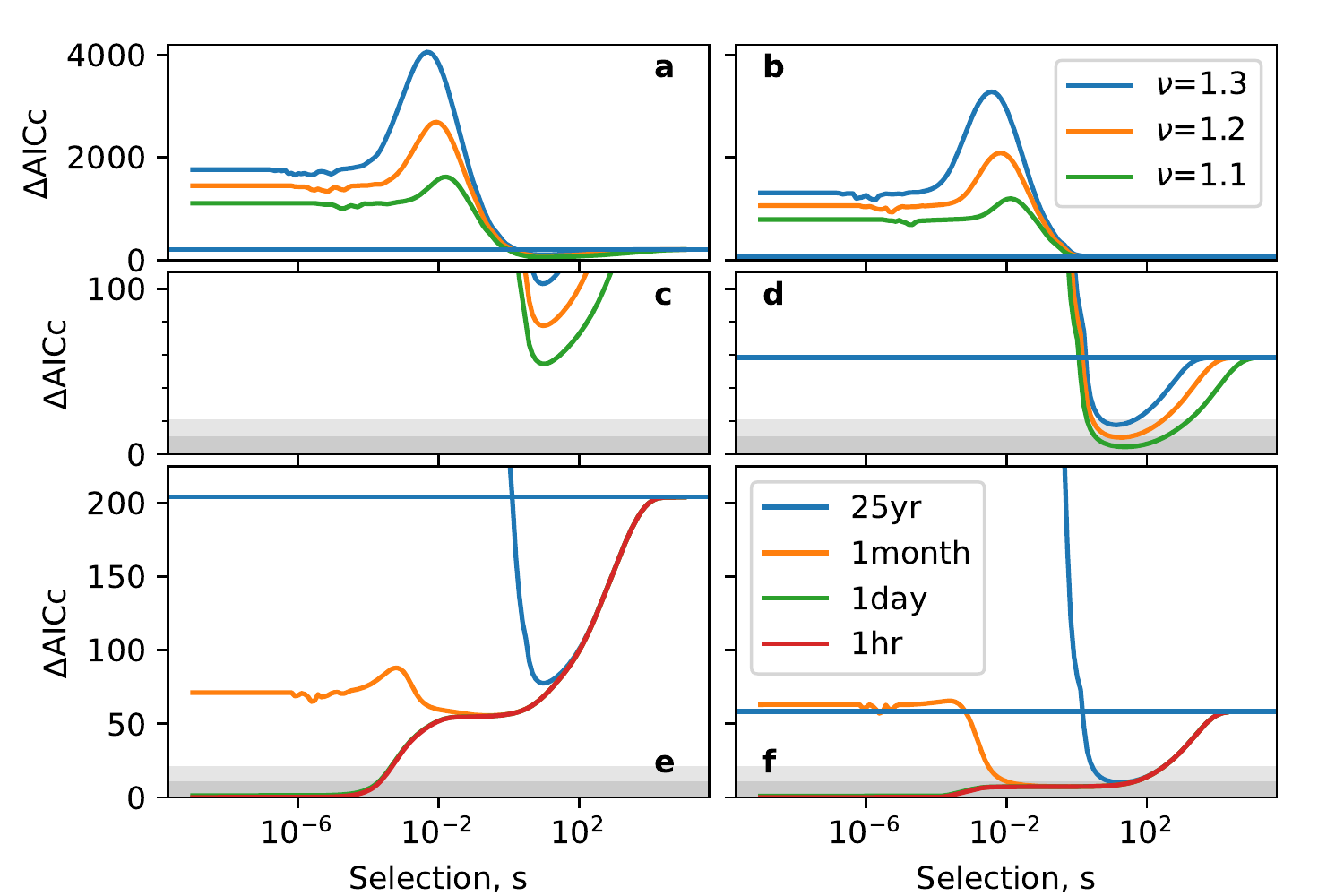} 
\caption{$\Daicc$ for models on heterogeneous social networks for the definite (panels a, c and e) and indefinite (b, d and f) articles as a function of selection strength $s$. Panels a--d show the effect of different degree exponents $\nu$ on the child-based model: panel c and d zoom in on $\Daicc\le100$, showing that plausibility is obtained only for the indefinite article over a limited range of $s$ and $\nu$. Panels e and f show the effect of memory lifetime at fixed $\nu=1.2$ and $\epsilon=1$. The horizontal line has the same meaning as in Fig.~\ref{fig:usage}. The dark and light shaded regions correspond to $\Daicc<10$ and $\Daicc<20$, respectively, which allows one to see the sensitivity to different evidence thresholds.}
\label{fig:network}
\end{figure}

In Fig~\ref{fig:network} we examine how the plausibility of both the child- and usage-based models investigated above changes when individual speakers in the model are arranged on complex network structures. This confirms our expectation that models in which timescales of change are less sensitive to population size receive greater support from the data.  As previously, the usage-based model provides a more plausible description of language change than the child-based model; moreover, the range of selection strengths and memory times over which a fit comparable to that provided by the Poisson process is much larger than on homogeneous networks.

We see from Fig~\ref{fig:network} that the most plausible models in the space under consideration are those in which selection is relatively weak. This is consistent with recent observations \cite{Newberry2017,Karjus2020,Karsdorp2020} that the dynamics of word frequencies appear to be subject to the evolutionary forces of both random drift and selection (i.e., neither is so strong that it dominates the other). Moreover, a number of studies (e.g., \cite{PastorSatorras2001,Lieberman2005,Antal2006}) have indicated that heterogeneity tends to lower the barrier to invasion of an infection, mutation or innovation. This possibly points towards a picture whereby the different grammatical structures that are attested cross-linguistically are somewhat similar in their fitness, but may nevertheless replace one another over time in the systematic way that is observed historically due to the manner in which human societies are structured.

\section{Discussion}

The aims of this work were twofold. First, we established how specific assumptions on the way in which individuals learn and use language translate to language change at the population scale. Second, we used historical data for the latter to identify which theories and mechanisms as to how individuals change the language of their speech community have greater empirical support.

Our main result is that if we impose the constraints that arise from assuming that childhood language learning is the driver of language change, there is no combination of the remaining free parameters that provides a good fit to the empirical data. The observed changes are many orders of magnitude more likely in regions of parameter space that correspond to other theories. The reason why the support for the child-based theory is so poor lies in a strong dependence of characteristic timescales at the population level on the underlying population size. If any selective bias in favour of the innovation is weak, the time taken for a change to propagate through a large speech community (the fixation time) is very much longer than the 100 years or so that is seen historically. If selection is strong, changes propagate quickly but then the rate at which successful changes are originated varies strongly with population size. The empirical data apparently show much less sensitivity to population size than the child-based theory implies.

In fact, throughout this work, we have found that the baseline model, which has no dependence on population size, fits the historical data well. One way to construe the baseline model is as changes originating once every 1000 years or so in every population, with changes then propagating rapidly through the population. This suggests that the mechanisms that have stronger empirical support are those that have these characteristics.

We acknowledge that our analysis is based on a single pair of features (the definite and indefinite articles) that are relatively unstable and are correlated. It is due to these correlations that we treated them separately (rather than combining them together into a single likelihood measure, which would assume independence). Nevertheless, comparison of the two articles is informative about how sensitive the analysis is to the details of which languages undergo a specific sequence of changes, as this does vary between the two articles. Overall, we find that it is the overall rate of language change combined with its weak sensitivity to population size that most strongly determines the plausibility of a given individual-based theory.

It is, however, possible that the dynamics of articles are unrepresentative of grammatical features more generally, and that our conclusions therefore do not generalise. We argue that this is unlikely. Regarding overall timescales of change, it is well established, by different analyses \cite{Wichmann2009,Dediu2010,Kauhanen2018}, that articles rank amongst the least stable of grammatical features and that others change more slowly. Basic word order lies at the opposite end of the spectrum, and the lifetime of given word orders have been estimated as ranging from $1000$--$100000$ years \cite{Maurits2014}. That is, these most stable structures persist for a timescale that ranges from around the same order of magnitude as articles to two orders of magnitude longer. A quick way to estimate the plausibility of the child-based theory for basic word order from our findings for articles is to consider a generational turnover that is increased by two orders of magnitude (i.e., from $25$ years to around $3$ months). Here we find a plausible account is possible on sufficiently heterogeneous social networks (see Fig.~\ref{fig:network}). This implies that the child-based theory could, at best, account for only the most stable grammatical structures, and does not offer a single explanation for language change that applies across the stability spectrum. The rate of population turnover imposes a fundamental minimum rate of language change which lies above that for unstable features in the child-based account, but potentially below in the usage-based account. Therefore the latter is capable of providing a common explanation for changes across the full stability spectrum.

It is harder to establish whether the weak sensitivity to population size is a feature of other grammatical changes. A detailed record of the history of each feature of interest across many languages is required for a conclusive assessment, data that is difficult to obtain (particularly for more stable features, where greater time depth is required to see a sufficiently large number of changes). However, a number of studies that have directly examined the relationship between population size and various aspects of language structure or change \cite{Wichmann2008,Lupyan2010,Nettle2012,Bromham2015} have tended to conclude that where there is an effect, it is weak. For example, \cite{Bromham2015} reports rates of gain and loss that scale sublinearly with the population size, consistent with the behaviour of Wright-Fisher models on heterogeneous social networks. Moreover, the fact that different methods \cite{Wichmann2009,Dediu2010,Kauhanen2018} of characterising the stability of a feature with a single metric are broadly consistent suggests that they do not vary significantly over space and time. Indeed, Wichmann and Holman \cite{Wichmann2009} have argued that the notion of stability is intrinsic to a feature and does not vary geographically. Given these considerations, it seems reasonable to conclude that weak population-size dependence is a generic property of language change, and not peculiar to articles.

We have identified two individual-level mechanisms that may contribute towards such a weak effect of population size on the rate of grammatical change. The first of these is provided for by usage-based accounts of language change which allow individuals to modify their behaviour across their lifespan, not just in the childhood language acquisition period. With more opportunities for individual behaviour to change per unit time, these theories allow changes to propagate through large speech communities more quickly. If the bias towards the innovation (the selection strength, $s$) is close to zero and the innovation rate per interaction is also small, changes at the population scale can then occur at roughly the same rate in different speech communities. 

In addition to small selection and innovation rates, this mechanism further requires a short memory lifetime in comparison to the lifetime of an individual (days or less, depending on social network structure). Taken at face value, such memory lifetimes may be considered unreasonably short. Here, we advise caution. First, a short memory does not imply that individual speakers are continually changing their behaviour: individual speakers can remain constant in their behaviour for as long as those around them do. If innovations rarely propagate, then most speakers will be exposed to existing conventions and continue to adhere to them, even though during a period of change they may alter their behaviour relatively quickly, albeit in small increments. There is some evidence that such changes can occur in older speakers as well as younger speakers, for example, in a study of Montreal French \cite{Sankoff2007}. Meanwhile, research on priming \cite{Pickering2004,Feher2016} shows that individual linguistic utterances can affect a speaker's behaviour in interactions in the very short term before fading away. It would be worth understanding whether such effects could effect more permanent changes, for example, when a change is in progress in a speech community, as this might then imply a shorter \emph{effective} memory time at the individual level than intuition grounded in everyday experience suggests.

The second mechanism that can reduce the sensitivity of grammatical change to population size are social network effects. Specifically, heterogeneous networks, in which a small number of well-connected speakers interact with a large number of poorly-connected speakers, lead to an effective population size (and therewith a characteristic timescale for change) that increases sublinearly with population size. Since this heterogeneity is a feature of certain social networks (e.g., those relating to phone calls, movie collaborations and social media \cite{Albert2002,Clauset2009,Kwak2010}), it is reasonable to assume that this is a property of human social interactions more generally. It is interesting to note that sublinear relationships between rates of change and population sizes have been reported in other empirical studies of language change \cite{Wichmann2008,Bromham2015}. Heterogeneous social networks offer one possible explanation for this phenomenon. To investigate this possibility further, it would be interesting to obtain more concrete information about the structure of linguistic interactions as well as how these stratify by age. If it were found, for example, that children's networks are more homogeneous than adult's, then this would point towards adults playing a key role in propagating an innovation throughout the speech community.

Although our statements about the relationship between individual behaviour and population-level change are grounded in a specific model of individual behaviour, we do not expect them to change if a different model was used. The reason for this is that any model that involves individual agents basing some or all of their future behaviour on that displayed by others (whether through learning or use) is expected to fall into the Wright-Fisher class \cite{Nordborg2019}. The precise relationship between parameter values in the individual-based model and those in the population-level origin-fixation model may vary between models: however, in any two models with similar memory lifetimes, innovation biases and social network structures would be expected to have the same behaviour at the population scale. In Appendix~\ref{app:demo}, we demonstrate this in the case of an extended model in which all properties vary between speakers, in which there is turnover in the population and social networks change over time.

This is not intended to imply that every feasible influence on language change is contained within the Wright-Fisher model used here (at least, at some level of abstraction). For example, we have excluded the possibility of a conformity bias \cite{Efferson2008,Eriksson2009}, wherein speakers suppress minority variants in favour of those in the majority. Such a bias however makes it increasingly difficult for innovations to propagate as the population increases in size, and therefore would be expected to exacerbate the problems of sensitivity to population size.  We have also assumed that factors influencing individual linguistic behaviour are constant over space and time. Specifically, social factors like prestige effects have been excluded, and it would be interesting in future work to establish whether these lead more readily to plausible accounts of historical language change.

\newpage

\appendix

{\LARGE\textbf{Appendices: Details of data and methods} }

These appendices set out in detail the empirical data set on article grammaticalisation cycles that was used in the main text, and in particular how a population size for each language was estimated. We also provide the explicit mathematical expressions for the likelihood functions that were used in the analysis, and explain in detail how we match up the dynamics of the Wright-Fisher model (that applies at the individual level) to the origin-fixation model (that applies at the population level).

\section{Empirical data set}
\label{app:data}

Our empirical data set of 52 languages is derived from 84 sources that document instances of article usage in different languages at different times. In Table~\ref{stab:histories} we summarise the stages of the grammaticalisation cycles for the definite and indefinite articles that have been observed, along with a historical time period that covers these observations. The quantities that enter the likelihood analysis are the total length of the historical period, and the number of changes in the article that occurred within it. In most cases, the beginning and end of each period corresponds to the earliest and most recent record (in cases where the language is still spoken, the latter is the present day). The exception to this is Hebrew, which was not spoken for a 1700-year period. In this case we take processes of change to be halted during this period. In cases where a language split into several daughter languages, we take the observation period for the parent language to end at the time of split, and the daughters' observation periods to begin. We also quote a measure of relative population size (the \emph{weight} of a language), which can be converted into an estimate of the true population size using the procedure described in Appendix~\ref{app:popsize} below. These weights are obtained from geographical population sizes as described in Appendix~\ref{app:langsize} below. Finally in this table we record the relevant sources of historical language use so our characterisation of the historical data can be verified as required.

\begin{longtable}{l|l|l|l|l|l}
Language & Period & Definite & Indefinite & Weight & References \\\hline
\endhead
English & 700CE -- 2000CE & 1,2 & 0,1,2 & $90.9$ & \cite{Traugott1992,Fischer1992} \\
German & 200CE -- 2000CE & 0,1,2 & 0,1 & $286$ & \cite{Harbert2007,Keller1978} \\
Common Scandinavian & 1100CE -- 1300CE & 3 & 0 & $61.3$ & \cite{Haugen1982} \\
Icelandic & 1300CE -- 2000CE & 3 & 0 & $1.00$ & \cite{Haugen1982} \\
Swedish & 1300CE -- 2000CE & 3 & 0,1 & $29.3$ & \cite{Haugen1982,Holmes1994} \\
Irish & 800CE -- 1600CE & 2 & 0 & $26.2$ & \cite{Thurneysen1946,Dillon1961} \\
Welsh & 1100CE -- 2000CE & 2 & 0 & $1.85$ & \cite{Evans1976,King2003} \\
Greek & 850BCE -- 2000CE & 1,2 & 0,1 & $55.0$ & \cite{Goodwin1892,Smyth1920,Horrocks2010,Holton1997} \\
Latin & 100BCE -- 500CE & 0,1 & 0 & $948$ & \cite{Clackson2007,Price1971,Maiden1995} \\
French & 500CE -- 2000CE & 1,2 & 0,1 & $366$ & \cite{Clackson2007,Price1971,Bourciez1956} \\
Romanian & 500CE -- 2000CE & 1,2,3 & 0,1 & $53.5$ & \cite{Bourciez1956} \\
Bulgarian & 800CE -- 2000CE & 0,1,2,3 & 0 & $32.8$ & \cite{Huntley1993,Scatton1993} \\
Russian & 800CE -- 2000CE & 0 & 0 & $323$ & \cite{Timberlake2004} \\
Egyptian & 5000BCE -- 700CE & 2,3 & 0,1 & $181$ & \cite{Loprieno1995,Allen2010,Junge2005} \\
Arabic & 700CE -- 2000CE & 3 & 3,0 & $458$ & \cite{Holes1995} \\
Hebrew & 1200BCE -- 200CE &  &  &  &  \\
 & 1900CE -- 2000CE & 3 & 0 & $20.5$ & \cite{Lambdin1971,Glinert1989,Coffin2005} \\
Persian & 500BCE -- 2000CE & 0 & 0,1,2,3 & $130$ & \cite{Skjaervo2005,Skjaervo2009,Lazard1992} \\
Indo Aryan & 700CE -- 1000CE & 0 & 0 & $2.51\times10^{3}$ & \cite{Masica1991} \\
Bengali & 1000CE -- 2000CE & 0,1,2,3 & 0,1,2,3 & $590$ & \cite{Masica1991} \\
Assamese & 1000CE -- 2000CE & 0,1,2,3 & 0,1,2,3 & $88.5$ & \cite{Masica1991} \\
Hindi & 1000CE -- 2000CE & 0 & 0 & $737$ & \cite{Masica1991} \\
Gujarati & 1000CE -- 2000CE & 0 & 0 & $118$ & \cite{Masica1991} \\
Korean & 900CE -- 2000CE & 0 & 0 & $82.1$ & \cite{Sohn1999} \\
Japanese & 700CE -- 2000CE & 0 & 0 & $231$ & \cite{Frellesvig2010} \\
Chinese & 1000BCE -- 2000CE & 0 & 0 & $3.39\times10^{3}$ & \cite{Norman1988} \\
Classical Nahuatl & 1500CE -- 1600CE & 2 & 1 & $22.1$ & \cite{Launey1981,Sullivan1988} \\
Tetelcingo Nahuatl & 1600CE -- 2000CE & 2,3,0 & 1 & $5.53\times10^{-3}$ & \cite{Tuggy1979} \\
North Pueblo Nahuatl & 1600CE -- 2000CE & 2 & 1,2 & $0.111$ & \cite{Brockway1979} \\
Michoacan Nahuatl & 1600CE -- 2000CE & 2,3,0,1 & 1 & $3.32\times10^{-3}$ & \cite{Beller1979} \\
Huasteca Nahuatl & 1600CE -- 2000CE & 2,3,0,1 & 1 & $1.31$ & \cite{Sischo1979} \\
Yucatec Maya & 1450CE -- 2000CE & 1 & 0,1 & $0.918$ & \cite{McQuown1967,Bolles1996} \\
Quiche Maya & 1400CE -- 2000CE & 1 & 1 & $2.15$ & \cite{Edmonson1967,LopezIxcoy1997} \\
Cakchiquel Maya & 1500CE -- 2000CE & 1,2 & 1 & $1.15$ & \cite{Maxwell2006,Brown2006} \\
Georgian & 300CE -- 2000CE & 1,2,3,0 & 0 & $4.81$ & \cite{Faehnrich1991,Tuite2004,Hewitt1995} \\
Armenian & 400CE -- 2000CE & 2,3 & 0,1,2 & $7.21$ & \cite{Clackson2004,DumTragut2009} \\
Aramaic & 950BCE -- 700CE & 3,0 & 0 & $28.9$ & \cite{Creason2004} \\
Geez & 400CE -- 1000CE & 0 & 0 & $12.1$ & \cite{Lambdin1978,Gragg1997} \\
Tigrinya & 1000CE -- 2000CE & 0,1 & 0 & $10.6$ & \cite{Faber1997,Kogan1997} \\
Tigre & 1000CE -- 2000CE & 0,1,2 & 0,1 & $1.51$ & \cite{Raz1983} \\
Akkadian & 2000BCE -- 600BCE & 0 & 0 & $43.7$ & \cite{Huehnergard2004,Moscati1980} \\
Sumerian & 3200BCE -- 2000BCE & 0 & 0 & $43.7$ & \cite{Michalowski2004,Jagersma2010} \\
Tamil & 250BCE -- 2000CE & 0 & 0,1 & $184$ & \cite{Steever2004,Rajam1992,Asher1982_1985,Caldwell1875_1961} \\
Tibetan & 800CE -- 2000CE & 0 & 0 & $13.5$ & \cite{Beyer1992,DeLancey2003,Denwood1999} \\
Mongolian & 1250CE -- 2000CE & 0 & 0 & $15.4$ & \cite{Poppe1974,Binnick1979} \\
Turkish & 1200CE -- 2000CE & 0 & 1 & $173$ & \cite{Kerslake1998,Lewis1967} \\
Khmer & 800CE -- 2000CE & 0 & 0 & $32.4$ & \cite{Sidwell2009,Jenner2010,Maspero1915} \\
Colonial Quechua & 1600CE -- 1700CE & 0 & 0 & $42.7$ & \cite{Adelaar2004} \\
Ayacucho Quechua & 1700CE -- 2000CE & 0 & 0 & $2.59$ & \cite{Adelaar2004,Parker1968} \\
Imbabura Quechua & 1700CE -- 2000CE & 0 & 0 & $0.683$ & \cite{Adelaar2004,Cole1985} \\
Huallaga Quechua & 1700CE -- 2000CE & 0 & 0 & $0.0399$ & \cite{Adelaar2004,Weber1989} \\
Aymara & 1600CE -- 2000CE & 0 & 0 & $8.55$ & \cite{Adelaar2004,Bertonio1603,Hardman2001} \\
Mapuche & 1600CE -- 2000CE & 0 & 1 & $17.8$ & \cite{Adelaar2004,Valdivia1060,Smeets2008} \\
\caption{\label{stab:histories}
Empirical dataset of historical language changes. Period: the historical periods
over which observations were made. Definite, Indefinite: Stages of the
grammaticalisation cycles observed for each article. Weight: relative historical
average population size.
    }
\end{longtable}

\section{Historical populations of geographical regions}
\label{app:popsize}

We use a survey of population sizes across many regions of the world \cite{McEvedy1978} to fit the following model for the size $N_i(t)$ of region $i$ at time $t$, measured in years since $1$BCE:
\begin{equation}
\label{popsi}
N_i(t) = w_i N_0 g(t) \;.
\end{equation}
In this model, $N_0$ sets an overall scale, $w_i$ is a region-dependent \emph{weight} that specifies its relative size, and $g(t)$ is a universal time-dependent growth function. This model amounts to an assumption that populations in different regions of the world maintain constant ratios over the relevant historical time period. Below we show that this model provides an estimate of a region's population size that is accurate to within a factor of 2.8 at a confidence level of $95\%$. The source data for this analysis is provided in the S2 File as an MS Excel spreadsheet; the resulting weights are specified in Table~\ref{stab:regwts}. Although not a particularly precise estimate, this variation of a factor of 2.8 is to be compared with a factor of 3400 variation in the weights themselves, and an overall increase by a factor of 300 in population sizes from 5000BCE to the present day. Consequently this simple model captures the range of population sizes and their changes over time rather well with a single parameter per geographical region, and we do not feel there would be much to be gained from using a more refined model.

\begin{longtable}{ll|ll}
Region & Weight & Region & Weight \\\hline
\endhead
Ancient Egypt & $181$ & Ireland & $26.2$ \\
Arabia & $105$ & Italy & $288$ \\
Austria & $41.2$ & Japan & $231$ \\
Bolivia & $18.3$ & Khmer Republic & $32.4$ \\
Bulgaria & $32.8$ & Korea & $82.1$ \\
C Turkestan Tibet & $53.9$ & Libya & $12.1$ \\
Caucasia & $40.1$ & Maghreb & $125$ \\
Chile & $17.8$ & Mexico & $111$ \\
China & $3.39\times10^{3}$ & Mongolia & $11.4$ \\
Czechoslovakia & $77.6$ & Nepal & $56.0$ \\
Denmark & $18.5$ & Norway & $13.5$ \\
Ecuador & $13.7$ & Pakistan India Bangladesh & $2.95\times10^{3}$ \\
Egypt & $114$ & Palestine Jordan & $20.5$ \\
England Wales & $92.7$ & Peru & $39.9$ \\
Ethiopia & $50.3$ & Poland & $84.7$ \\
France & $366$ & Romania & $53.5$ \\
Germany & $245$ & Russia In Europe & $323$ \\
Greece & $55.0$ & Sri Lanka & $32.0$ \\
Guatemala & $14.3$ & Sweden & $29.3$ \\
Iberia & $241$ & Syria Lebanon & $38.5$ \\
Iceland & $1.00$ & Turkey In Asia & $230$ \\
Iran & $130$ & Yugoslavia & $81.5$ \\
Iraq & $43.7$ &  &  \\
\caption{\label{stab:regwts}
Relative historical average size of each geographical region relevant to the
languages in the sample (all to 3~s.f.). These sizes are normalized such that
the smallest such region (Iceland) has a relative size of 1.
    }
\end{longtable}

\begin{figure}
\begin{center}
\includegraphics[width=0.6\linewidth]{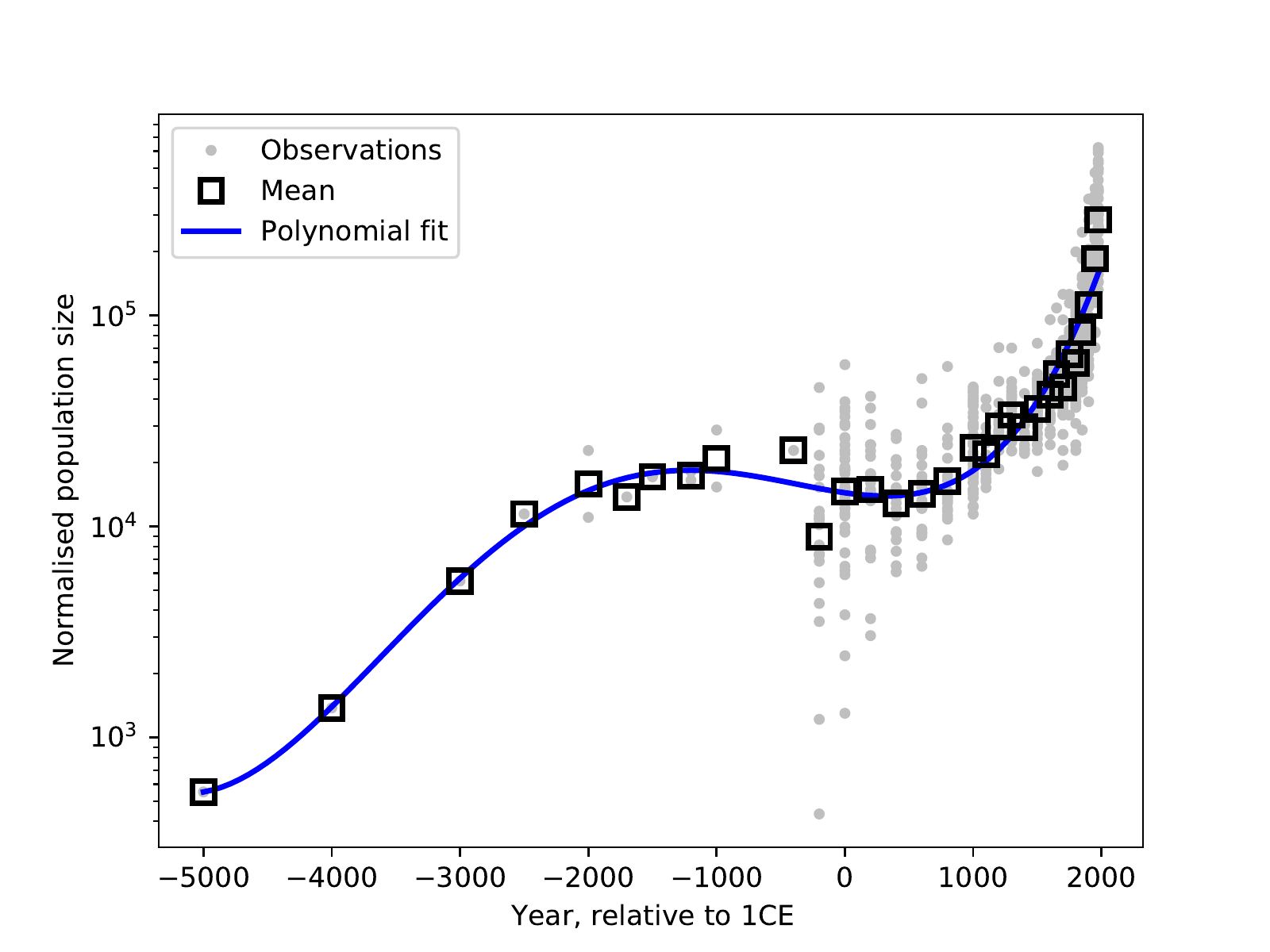}
\end{center}
\caption{\label{sfig:popsizes} Population sizes normalised by by the weight factor $w_i$ for each geographical region $i$. The grey dots show observations recorded in \cite{McEvedy1978}. The black squares are the mean population size after normalisation has been applied to minimise the variance between different regions. The smooth line is a degree 4 polynomial fit to the logarithm of the mean normalised population size. See Table~\ref{stab:polycoeffs} for the coefficients in this polynomial.}
\end{figure}

\begin{figure}
\begin{center}
\includegraphics[width=0.6\linewidth]{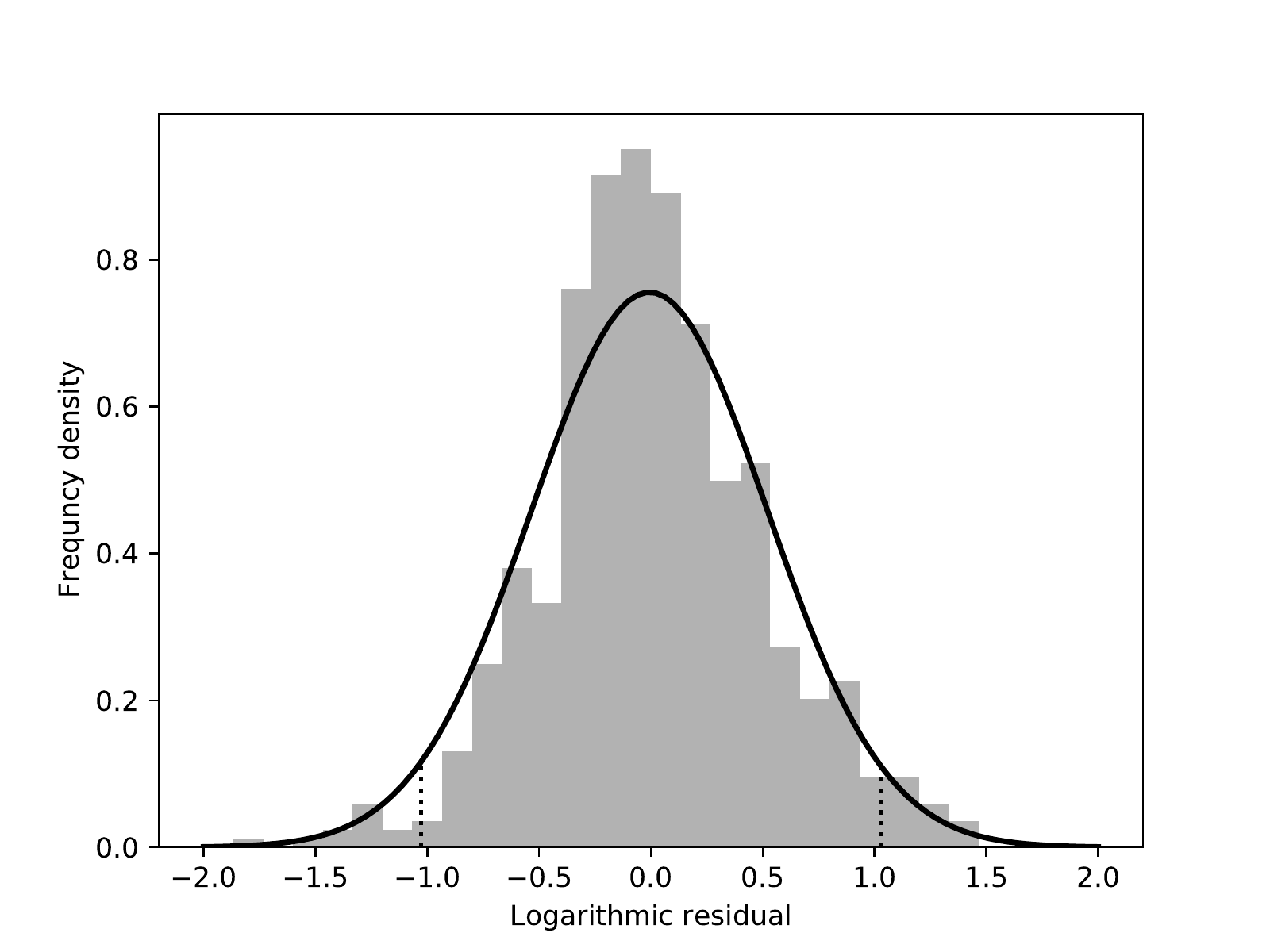}
\end{center}
\caption{\label{sfig:residuals} Distribution of the logarithmic residuals, $\ln (N_{\rm obs}/N_{\rm fit})$ where $N_{\rm obs}$ and $N_{\rm fit}$ are the observed and fit population sizes, respectively. Solid line: normal distribution with the same mean and variance. Dashed lines: extent of the central 95\% of the residuals.}
\end{figure}

The weights and the unknown function $g(t)$ are found by performing a linear least-squares fit to
\begin{equation}
\label{lnpopsi}
\ln(N_i(t_j)) = \ln(w_i) + \ln(N_0) + \ln(g(t_j)) + \epsilon_i(t_j) \;,
\end{equation}
where $N_i(t_j)$ is the population size in region $i$ at time point $t_j$ as recorded in \cite{McEvedy1978}, and the parameters $a_i=\ln(w_i)$ and $b_j=\ln(g(t_j))$ are varied to minimise the sum of square residuals $\epsilon_i(t_j)$. This minimisation problem is underdetermined, and a unique solution is obtained after fixing $g(0)=1$ and $w_i=1$ for the smallest region in the sample (Iceland). This procedure yields an overall scale $N_0 = 14600$ (to 3 s.f.), and the weights $w_i$ are presented in Table~\ref{stab:regwts}. In Figure~\ref{sfig:popsizes}, we plot the normalised population sizes $N_i(t_j)/w_i$, along with a fit to its mean, $g(t)$, whose logarithm is found to be well described by a quartic polynomial. This figure demonstrates that there is some scatter around this average, which we quantify further with the distribution of residuals that is shown in Figure~\ref{sfig:residuals}. We find that although this distribution is not normal, the central $95\%$ of the residuals span the interval from  $-1.02$ to $1.3$. Since these residuals are natural logarithms, this range corresponds to the overall factor of $2.8$ error (in either direction) on the estimated population size, as claimed above.  The $R^2$ statistic for the linear least-squares fit is $0.923$ (to 3~s.f.). 

\begin{longtable}{l|l|l|l|l|l}
Coefficient & $c_{0}$ & $c_{1}$ & $c_{2}$ & $c_{3}$ & $c_{4}$ \\\hline
\endhead
Value & $-0.0127$ & $-2.00\times10^{-4}$ & $2.13\times10^{-7}$ & $2.04\times10^{-10}$ & $2.55\times10^{-14}$ \\
\caption{\label{stab:polycoeffs}
Coefficients in the polynomial fit $c_0 + c_1 t + c_2 t^2 + c_3 t^3 + c_4 t^4$
to the function $\ln (g(t))$ in (\ref{lnpopsi}) obtained by least-squares minimisation.
    }
\end{longtable}

\clearpage

\section{Geographical composition of languages}
\label{app:langsize}

We assume that language sizes are linear combinations of the sizes of the geographical regions where they are spoken, and hence that these fractions remain constant over time. These combinations are provided in Table~\ref{stab:commcomps}, with the resulting language sizes in Table~\ref{stab:histories}. For most cases, we assume a one-to-one relationship between a geographical region (e.g., Sweden) and a language (Swedish). In other cases, further explanation is necessary:
\begin{itemize}
\item \textit{English and Welsh} --- Currently, around 5\% of the population of England and Wales resides in Wales. Assuming this fraction to be constant over history, and that roughly one third of Welsh residents are speakers of Welsh, we arrive at a 98\% to 2\% split across England and Wales into English and Welsh speakers respectively.
\item \textit{Scandinavian} --- Two of the languages in the sample, Icelandic and Swedish, descended from Common Scandinavian with the split estimated to have occurred around 1300CE. We therefore take the period from 1100CE to 1300CE as common to both languages, and to have a correspondingly larger population (see below for details of how this was incorporated into the analysis).
\item \textit{Irish} --- Until around 1600CE, we assume the entire population of Ireland is Irish-speaking before the Irish-speaking population declines. We therefore truncate the time-window for Irish in the analysis at 1600CE, rather than the present day (see Table~\ref{stab:histories}).
\item \textit{Latin} --- Analogously to the Scandinavian languages, we take Latin to be an ancestor of French and Romanian, estimating the split to take place around 500CE, before which both daughter languages are considered to have a common history.
\item \textit{Turkish} --- The region designated Turkey-in-Asia by McEvedy and Jones \cite{McEvedy1978} includes a significant population of Kurdish, Armenian and Greek speakers. We take the number of Turkish speakers to be 75\% of this larger population.
\item \textit{Hebrew} --- Hebrew was not spoken between around 200CE and 1900CE, being a liturgical and written language in the intervening period. Here, we assume that the language was frozen (unable to change) in the period that it was not spoken.
\item \textit{Aramaic} --- McEvedy and Jones \cite{McEvedy1978} combine Syria (where Aramaic was spoken) and Lebanon into a single region. We estimate that Aramaic speakers make up 75\% of this region.
\item \textit{Ethiopian Semitic} --- Ge`ez is assumed to be the direct ancestor of both Tigr\'e and Tigrinya, with a split occouring at 1000CE. The fraction of Ethiopia in which each daughter language is spoken is assumed to be constant and equal to 1986 values taken from \cite{Grimes1988}.
\item \textit{Indo-Aryan} --- Masica \cite[p8]{Masica1991} quotes a figure of 640m total speakers of Indo-Aryan languages. The majority of these reside in the region designated as Pakistan, India and Bangladesh by McEvedy and Jones \cite{McEvedy1978}, for which the most recent estimate of population size (1975CE) was given as 745m. We therefore assume the population of Indo-Aryan speakers to track the size of Pakistan, India and Bangladesh, but scaled by a factor of $85\%$ to match recent estimates in \cite{Grimes1988}. We take the Indo-Aryan ancestor language to split at around 1000CE; the daughter languages Bengali, Assamese, Hindi and Gujarati that were included in the sample are thereby taken to share a common period of evolution from around 700CE to 1000CE (as with the Scandinavian and Latin languages). These sizes of each of these daughter languages were also taken to track the population of Pakistan, India and Bangladesh, again using the ratio that applies to the present-day populations.
\item \textit{Tamil} --- Tamil is spoken in India and Sri Lanka. Proportions of the corresponding regions documented by McEvedy and Jones \cite{McEvedy1978} were estimated using numbers for 1986CE in \cite{Grimes1988}.
\item \textit{Tibetan} --- We take 25\% of the population of the Chinese Turkestan and Tibet region \cite{McEvedy1978} to be Tibetan.
\item \textit{Mongolian} --- The Mongolian-speaking population extends beyond the region designated as Mongolia by McEvedy and Jones which excludes Inner Mongolia. We assume that the total population of Mongolian speakers is 35\% larger than that of Mongolia.
\item \textit{Mesoamerican languages} --- These languages are spoken in regions designated as Mexico and Guatemala by McEvedy and Jones \cite{McEvedy1978}. We take current estimates of their speaker numbers as fractions of the relevant geographical regions to obtain the weight of the languages, and a historical estimate to determine the fraction of Mexico population that spoke Classical Nahuatl before splitting into daughters at around 1600CE. However, given the recent decline (in particular) of Nahuatl as it was replaced by Spanish in Mexico, this means that the numbers post-split are likely to be underestimates, and furthermore it may not be reasonable to assume that the speaker numbers are a constant fraction of the geographical population over time. However, we do not believe that this uncertainty greatly affects our results.
\item \textit{Quechua} --- We consider three varieties of Quechua, all of which are taken to be descendants of a common Colonial Quechua language spoken widely across Bolivia, Ecuador and Peru before 1700CE. After this split, we assume that the speakers of the daughter varieties are constant fractions of Peru and Ecuador set at values that pertain to the 1970s \cite{Grimes1988}. 
\end{itemize}

\begin{longtable}{l|l}
Language & Geographical composition \\\hline
\endhead
Akkadian & 100\% Iraq \\
Arabic & 100\% Arabia + 100\% Iraq + 100\% Palestine Jordan + 100\% Syria Lebanon \\
 & \hspace{1em} + 100\% Maghreb + 100\% Libya + 100\% Egypt \\
Aramaic & 75\% Syria Lebanon \\
Armenian & 18\% Caucasia \\
Assamese & 3\% Pakistan India Bangladesh \\
Ayacucho Quechua & 6.5\% Peru \\
Aymara & 25\% Bolivia + 10\% Peru \\
Bengali & 20\% Pakistan India Bangladesh \\
Bulgarian & 100\% Bulgaria \\
Cakchiquel Maya & 8\% Guatemala \\
Chinese & 100\% China \\
Classical Nahuatl & 20\% Mexico \\
Colonial Quechua & 50\% Bolivia + 100\% Ecuador + 50\% Peru \\
Common Scandinavian & 100\% Denmark + 100\% Sweden + 100\% Norway \\
Egyptian & 100\% Ancient Egypt \\
English & 98\% England Wales \\
French & 100\% France \\
Geez & 24\% Ethiopia \\
Georgian & 12\% Caucasia \\
German & 100\% Germany + 100\% Austria \\
Greek & 100\% Greece \\
Gujarati & 4\% Pakistan India Bangladesh \\
Hebrew & 100\% Palestine Jordan \\
Hindi & 25\% Pakistan India Bangladesh \\
Huallaga Quechua & 0.1\% Peru \\
Huasteca Nahuatl & 1.18\% Mexico \\
Icelandic & 100\% Iceland \\
Imbabura Quechua & 5\% Ecuador \\
Indo Aryan & 85\% Pakistan India Bangladesh \\
Irish & 100\% Ireland \\
Japanese & 100\% Japan \\
Khmer & 100\% Khmer Republic \\
Korean & 100\% Korea \\
Latin & 100\% France + 100\% Iberia + 100\% Italy + 100\% Romania \\
Mapuche & 100\% Chile \\
Michoacan Nahuatl & 0.003\% Mexico \\
Mongolian & 135\% Mongolia \\
North Pueblo Nahuatl & 0.1\% Mexico \\
Persian & 100\% Iran \\
Quiche Maya & 15\% Guatemala \\
Romanian & 100\% Romania \\
Russian & 100\% Russia In Europe \\
Sumerian & 100\% Iraq \\
Swedish & 100\% Sweden \\
Tamil & 6\% Pakistan India Bangladesh + 23\% Sri Lanka \\
Tetelcingo Nahuatl & 0.005\% Mexico \\
Tibetan & 25\% C Turkestan Tibet \\
Tigre & 3\% Ethiopia \\
Tigrinya & 21\% Ethiopia \\
Turkish & 75\% Turkey In Asia \\
Welsh & 2\% England Wales \\
Yucatec Maya & 0.83\% Mexico \\
\caption{\label{stab:commcomps}
Geographical composition of the speech community for each language in the sample.
The resulting relative historical average population sizes are given in
Table~\ref{stab:histories}.
}
\end{longtable}

In the main text, the model calls for a single size to characterise each population over the relevant historical period. We use the mean of $N_i(t)$ given by Eq.~(\ref{popsi}) over the historical time period (or periods, for Hebrew) given in Table~\ref{stab:histories}, summed over regions $i$ with the weights given in Table~\ref{stab:commcomps}.

\section{Origin-fixation model}
\label{app:ofmodel}

As described in the main text, we have constructed an origin-fixation model to describe the dynamics of language change at the population level, and more specifically provide likelihood functions for the empirical data sets under various assumptions on the underlying linguistic behaviour of individuals. In this model, an innovation (mutation) of type $i+1$ that successfully propagates through (invades) a population of type $i$ individuals is introduced as a Poisson process with rate $\omega_i$. Recall from the main text that we refer to the introduction of a successful innovation to the population as an \emph{origination} event. We take $\omega_i = \bar{\omega}/{(Vf_i)}$, where $f_i$ is the typological frequency of variant $i=1,2,\ldots,V$ across the world's languages, so that mean time between the onset of origination events is proportional to $f_i$.  In most origin-fixation models \cite{McCandlish2014}, the fixation time $T_F$ is idealised as zero. As explained in the main text, in language change the timescale of fixation (decades to hundreds of years) is not greatly separated from the origination timescale (hundreds to thousands of years). Consequently, we must generalise to nonzero fixation times.  Specifically, we model the fixation process as one whose time to fixation is drawn from a Gamma distribution (this because the fixation time is necessarily positive, and we wish to treat its mean and variance as independent quantities). Since in this framework it is possible that the change to the next stage of the cycle may be triggered before the previous one has gone to fixation, we must also account for interference between successive originations.

\subsection{Likelihood function}

In the main text, we compared different models using the Akaike Information Criterion, defined through equation (4). This involves the
likelihood function
\begin{equation}
{\cal L} = \prod_{i=1}^{n} {\cal L}_{m_i}(t_i)
\end{equation}
where ${\cal L}_{m_i}(t_i)$ gives the probability that exactly $m_i$ language changes have occurred in a time window of length $t_i$ these corresponding to language $i$ in the sample. In this section, we explain how ${\cal L}_{m_i}(t_i)$, and therewith ${\cal L}$, is calculated.

We assume that at the beginning of an observation window, $t=0$, only one variant is present in the population. Let $\omega_1$ be the origination rate at the first stage in the cycle and $\overline{T_F}$ and $\sigma_{T_F}^2$ be the mean and variance of the time for the innovation to reach fixation in the population, conditioned on this event occurring. Even in those cases where exact results are available \cite{Crow1970}, the functional form of the distribution $p_1(t)$ that the invading mutant fixes at time $t$ is very complicated. We have found it is well approximated by the convolution of a Poisson process with rate $\omega$ and a Gamma distribution with mean $\overline{T_F}$ and variance $\sigma_{T_F}^2$ (see Section~\ref{app:test} below). That is
\begin{equation}
p_1(t) \approx \omega_1 {\rm e}^{-\omega_1 t} \ast \frac{(\beta t)^{\alpha-1} \beta {\rm e}^{-\beta t}}{\Gamma(\alpha)}
\end{equation}
where $\Gamma(\alpha)$ is the Gamma function and $\ast$ denotes the convolution operation. The parameters $\alpha$ and $\beta$ are related to the mean and variance of the fixation time via
\begin{equation}
\overline{T_F} = \frac{\alpha}{\beta} \quad\mbox{and}\quad \sigma_{T_F}^2 = \frac{\alpha}{\beta^2} \;.
\end{equation}

Once the innovation has gone to fixation, the next innovation can then be introduced to the population at rate $\omega_2$ (which need not equal $\omega_1$) and its fixation time is assumed to have the same mean and variance as the first mutant. The probability $P_m(t)$ that \emph{at least} $m$ changes have occurred by time $t$ is obtained by convolving $p_1(t)$ with itself $m$ times, and integrating from $0$ to $t$. The probability ${\cal L}_m(t)$ that \emph{exactly} $m$ changes have occurred by time $t$ is then given by $\int_0^t [ P_m(t')-P_{m+1}(t')] \,{\rm d}t'$. This is most conveniently written in the form of the Laplace transform
\begin{equation}
\label{LT}
\hat{\cal L}_m(s) = \frac{1}{s} \left( \prod_{i=1}^{m} \frac{w_i}{w_i+s} \right) \left( \frac{\beta}{\beta+s} \right)^{m\alpha} \left[ 1 - \frac{w_{m+1}}{w_{m+1}+s} \left( \frac{\beta}{\beta+s} \right)^{\alpha} \right] \;.
\end{equation}
Note that in the case $m=0$, the product over $i$ is set equal to unity.

It is possible to invert the Laplace transform and obtain explicit expressions for ${\cal L}_m(t)$ involving alternating sums of incomplete Gamma functions. Unfortunately, these expressions are difficult to compute to the desired numerical precision, due to  cancellations between terms at the leading order. A much better approach is to numerically invert (\ref{LT}) using the Euler algorithm as set out in \cite{Abate2006}. This involves computing the sum
\begin{equation}
\label{Linv}
{\cal L}(t) \approx \frac{1}{t} \sum_{i=1}^{n} c_i \Re \hat{\cal L}_m\left(\frac{\zeta_i}{t}\right)
\end{equation}
where the number of terms $n$, the weights $c_i$ and the nodes $\zeta_i$ depend on the desired precision \cite{Abate2006}. We have found five digits of precision sufficient for our needs, which corresponds to $n=18$, a remarkably small number of function evaluations given the complexity of the problem.  The different cultural evolutionary scenarios that we test in the main text give rise to a wide range of different parameter combinations. To maintain numerical precision in calculating the logarithm of the likelihood ${\cal L}_m(t)$ across the full range of parameter values, a few minor modifications to the standard Euler algorithm were required:
\begin{itemize}
\item For the case $m=0$, ${\cal L}_0(t)\to 1$ as $t\to0$. Here, the log likelihood is close to zero, and therefore for this to be obtained to the desired precision, we  invert the transform of $1-{\cal L}_0(t)$ (which is close to zero), and use a library function to evaluate $\ln(1-x)$ for small $x$.
\item In all other cases, the likelihood has a leading exponential decay
\[ {\rm e}^{-s^\ast t} \quad\mbox{where}\quad s^\ast = \min\{ \beta, \omega_1, \ldots, \omega_m \}\]
is the location of the singularity in (\ref{LT}) closest to the origin in the complex-$s$ plane. Since the combination $s^\ast t$ can become large (leading to a very small likelihood), we maintain precision via the identity $\ln {\cal L}(t) = - s^\ast t + \ln {\cal R}_{s^\ast}(t)$ where ${\cal R}_{s^\ast}(t)$ is the inverse of the shifted Laplace transform $\hat{{\cal L}}(s-s^\ast)$. Note that the apparent pole at $s=0$ is cancelled by a zero in the numerator, which permits this shift of the integration contour.
\item Finally, some of the terms in the sum (\ref{Linv}) can take values that are sufficiently small or large to cause overflow when working to machine precision. We handle this by computing the logarithm of each term in the sum and subtracting out the largest real part from all terms. Then the remainders can be safely exponentiated and summed without causing overflow. The contribution that was subtracted is then reinstated into the result for the log likelihood at the end of the calculation.
\end{itemize}
The likelihood of the macroscopic Poisson process with state-dependent origination rates $\omega_1, \omega_2, \ldots$ can be obtained from the inversion of (\ref{LT}) after taking the limit $\beta\to\infty$. A complete implementation of the likelihood analysis code is provided for reference at \cite{code}. 

\subsection{Interference correction}
\label{app:interfere}

The calculation of the likelihood function above is conditioned on each origination going to fixation before the next origination event is triggered. When the origination rate $\omega_i$ is comparable to $1/\overline{T_F}$, successive origination events can interfere, which ultimately leads to coexistence of multiple variants rather than a sequence of changes going through the population. This feature is inconsistent with the empirical data, in which the typical situation is that one convention dominates, or that we are in transition from one stage of the cycle to the next.  To prevent a maximum of the likelihood function being found in this region, we need to multiply it by the probability that an originated innovation \emph{does} go to fixation: this then gives us the joint probability that fixation will occur, and that it has occurred by a given time. 

 We introduce therefore the correction $C_i$, which is the propbability that the innovation introduced at stage $i$ (i.e., the one that precipitates a change to stage $i+1$) goes to fixation without interference. Then,
\begin{equation}
{\cal L}_m'(t) = \left(\prod_{i=1}^{m} C_i\right) {\cal L}_m(t) \;.
\end{equation}
In the case of the Poisson process (or the Wright-Fisher model with infinite selection) $T_F\equiv0$, and so no correction is needed ($C_i \equiv 1$). In the general Wright-Fisher model, we find from numerical computations that $C_i$ is well approximated by $C_i = {\rm e}^{-\omega_i \overline{T_F}}$ (see Section~\ref{app:test} below). However, the precise form of the correction is not too important in terms of likelihood maximisation, as long as $C_i \approx 1$ when innovations can propagate freely without interference, and decreases towards $0$ when they cannot.

\section{Wright-Fisher model}
\label{app:wfmodel}

At the individual speaker level, we use a Wright-Fisher model. We provide the full definition of this model here, explain how we extract from it the parameters $\alpha$, $\beta$ and $\omega_i$ in the origin-fixation model, and demonstrate numerically that the latter serves as a good approximation to the Wright-Fisher model after averaging over individual innovation frequencies.

\subsection{Definition}

The dynamics of the Wright-Fisher model are illustrated in Fig.~1 of the main text. We assume that at a given point in time, the language is in transition from stage $i$ to $i+1$ of the cycle. Each of the $N$ speakers is then characterised by the frequency $x_n$ that they use the innovation (i.e., produce an utterance consistent with stage $i+1$ of the cycle).  Each speaker updates their grammar (their $x_n$ value) at intervals of $\Delta t = 1/R$, where $R$ is the interaction rate. They retain a fraction $(1-\epsilon)$ of their existing grammar, and replace the remaining fraction with a memory of either the innovation or the existing convention, this depending on the relevant linguistic interactions they have over the interval $\Delta t$. If we define a variable $\tau_n$ such that $\tau_n=0$ when the speaker stores a memory of the convention, and $\tau_n=1$ for the case of the innovation, we have
\begin{equation}
x_n' = (1-\epsilon) x_n + \epsilon \tau_n \;.
\end{equation}

The probability $p_n$ that $\tau_n=1$ depends on the frequency of the innovation in the speaker's neighbourhood, $\bar{x}_n$, the individual innovation rate $\eta$ and the selection strength $s$. The specific prescription is
\begin{equation}
p_n = \frac{1-\bar{x}_n}{1+\bar{x}_n s} \eta_i + \frac{(1+s)\bar{x}_n}{1+\bar{x}_n s} \;.
\end{equation}
In words, this equation says that a speaker first samples an instance of linguistic behaviour from their local neighbourhood, wherein the convention is given weight $1$, and the innovation weight $1+s$. That is, if $s>0$ the innovation is selected for; and if $s<0$ it is selected against. If this instance of linguistic behaviour corresponds to the convention, there is a probability $\eta_i$ that it is recorded by the speaker as the innovation (i.e., a mutation from stage $i$ to $i+1$ of the cycle has occurred at the individual level).

We are deliberately abstract in our specification of the model, as different mechanisms can give rise to selection and innovation. A number of concrete examples, and their relation to linguistic theories, are provided in the main text.

\subsection{Correspondence with the origin-fixation model}

To connect the Wright-Fisher model to the origin-fixation model, we need to work out the values of the parameters $\omega_i$, $\overline{T_F}$ and $\sigma_{T_F}^2 \equiv \overline{T_F^2} - \overline{T_F}^2$ of the latter that are implied by the former. We consider first of all the origination rates. Starting from a state with $x_n=0$ for all speakers, we see that each agent has a probability $\eta_i$ of generating an innovation at a rate $R$. Then the total rate at which successful innovations propagate is $N R \eta_i Q(\epsilon/N)$, where $Q(x_0)$ is the probability that an innovation with frequency $x_0$ in the \emph{population} goes to fixation. $x_0=\epsilon/N$ because we assume the innovation rate is sufficiently small that exactly one speaker starts off with the innovation at level $\epsilon$. Within this low innovation-rate regime, the innovation then propagates under the influence of selection and drift (fluctuations arising from the finite exposure to linguistic behaviour) until it reaches fixation. 

To calculate $Q(x_0)$, we first define $\delta x_n = x_n' - x_n$ and determine the expectation values $\langle \delta x_n \rangle$ and $\langle (\delta x_n)^2 \rangle$ over the distribution of $\tau_n$ given above. This allows one to write down the forward or backward Kolmogorov equation for the set of individual speaker frequencies $x_n$ via the Kramers-Moyal expansion. Various studies, e.g., \cite{Sood2005,Antal2006,Baxter2008}, have shown that for an appropriate weighted average $x$ of these individual speaker frequencies, one has the backward Kolmogorov equation
\begin{equation}
\label{bfpe}
T_M \frac{\partial q(t|x)}{\partial t} = s x(1-x) \frac{\partial q(t|x)}{\partial x} + \frac{1}{2N_e} x(1-x)\frac{\partial^2 q(t|x)}{\partial x^2} 
\end{equation}
for the probability distribution $q(t|x)$ that an innovation with initial frequency $x$ reaches fixation at time $t$. In this equation, $s$ is as defined in the Wright-Fisher model, $T_M = 1 / (R\epsilon)$ is the memory lifetime identified in the main text, and $N_e$ is an effective population size
\begin{equation}
N_e = \frac{N}{\epsilon} \frac{\overline{z}^2}{\overline{z^2}} \;.
\end{equation}
The quantity $z_n$ is defined as the number of others speakers that speaker $n$ can observe the linguistic behaviour of; we assume that each of these speakers is given the same amount of attention, although the number of neighbours $z_n$ can vary across a social network. (Derivations of this result can be found in \cite{Sood2005,Antal2006,Baxter2008}).

This Kolmogorov equation is extremely well studied in the population genetics literature (e.g.~\cite{Kimura1969,Crow1970}). In particular, the procedure for obtaining moments of the time to reach fixation from a single mutant ($x\to0$), conditioned on fixation occurring, is well established \cite{Kimura1969}. To compute $\overline{T_F}$ and $\sigma_{T_F}^2$, we require the first two moments.  Defining
\begin{equation}
F_k(x) = \int_0^{\infty}  t^k q(t|x) \, {\rm d} t \;,
\end{equation}
we have
\begin{equation}
\label{moments}
Q(x) = F_0(x) \;,\quad \overline{T_F} = \lim_{x\to0} \frac{F_1(x)}{F_0(x)} \quad\mbox{and}\quad
\overline{T_F^2} = \lim_{x\to0} \frac{F_2(x)}{F_0(x)}
\end{equation}
where $Q(x)$ is the probability that a mutant with initial frequency $x$ fixes.  By multiplying (\ref{bfpe}) by $t^k$ and integrating \cite{Kimura1969}, we find the recursion 
\begin{equation}
\label{fpde}
2N_e s x(1-x) F_k'(x) + x(1-x) F_k''(x) = - 2 k N_e T_M F_{k-1}(x)
\end{equation}
where a prime denotes differentiation. The boundary conditions of this differential equation are $F_k(0)=0$ for all $k$, $F_0(1)=1$ and $F_k(1)=0$ for $k>0$.  For the case $k=0$, the solution has the closed form  \cite{Kimura1969}
\begin{equation}
Q(x) = F_0(x) = \frac{1 - {\rm e}^{-2N_e s x}}{1 - {\rm e}^{-2N_e s}} \;.
\end{equation}
Substituting $x=\frac{\epsilon}{N}$ we arrive at Eq.~(3) in the main text. Although no general closed-form expression exists for the case $k>0$, (\ref{fpde}) can still be integrated to obtain \cite{Kimura1969}
\begin{equation}
\label{I1}
F_k(x) = 2 k N_e T_M \left[ (1 - Q(x)) \int_0^x \frac{F_{k-1}(y)}{y(1-y)} \frac{Q(y)}{Q'(y)} \, {\rm d} y + Q(x) \int_p^1 \frac{F_{k-1}(y)}{y(1-y)} \frac{1-Q(y)}{Q'(y)}  \,{\rm d} y  \right]
\end{equation}
which can be evaluated numerically and substituted into (\ref{moments}) to obtain the desired moments.

The complexity of this numerical problem is simplified slightly by noting that
\begin{equation}
\label{I2}
\overline{T_F^k} = \lim_{x\to0} \frac{F_k(x)}{F_0(x)} = 2 k N_e T_M \int_0^1  \frac{F_{k-1}(y)}{y(1-y)} \frac{1-Q(y)}{Q'(y)} \,{\rm d} y = \frac{k T_M}{s} \int_0^1  \frac{F_{k-1}(y)}{y(1-y)} \left[1- {\rm e}^{-2N_e s(1-y)}\right] {\rm d} y  
\end{equation}
and the fact that $\overline{T_F^k}$ is symmetric in $s \to -s$. However, we find that a numerical integration routine, implemented na\"{\i}vely, becomes unreliable for small and large $N_e |s|$. In these regimes it both more efficient and less susceptible to numerical instability to use Taylor series and asymptotic expansions, respectively. Specifically, for small $|s|$ we use the approximations
\begin{align}
\overline{T_F} &\approx 2 N_e T_M \left[ 1 - \frac{(2N_e s)^2}{72} \right] &\mbox{when $2N_e |s| < 10^{-3}$} \\
\overline{T_F^2} &\approx 8 N_e^2 T_M^2 \left[ \frac{\pi^2}{3}-2 + \left( \frac{\pi^2}{36} - \frac{17}{54} \right) (2 N_e s)^2 \right] &\mbox{when $2N_e |s| < 10^{-2}$}
\end{align}
and for large $|s|$
\begin{align}
\overline{T_F} &\sim \frac{2T_M}{|s|} \left[ \ln(2N_e |s|) + \gamma - \frac{1}{2N_e |s|} \right] &\mbox{when $2N_e |s| > 500$} \\
\overline{T_F^2} &\sim \frac{2T_M^2}{s^2} \left[ 4 (\ln(2N_e |s|) + \gamma)^2 + \frac{\pi^2}{3} \right] &\mbox{when $2N_e |s| > 500$}
\end{align}
in which $\gamma$ is the Euler-Mascheroni constant $\gamma = 0.577\ldots$.

\subsubsection{Numerical test of the correspondence between the models}
\label{app:test}

\begin{figure}
\centering
\includegraphics[width=.9\linewidth]{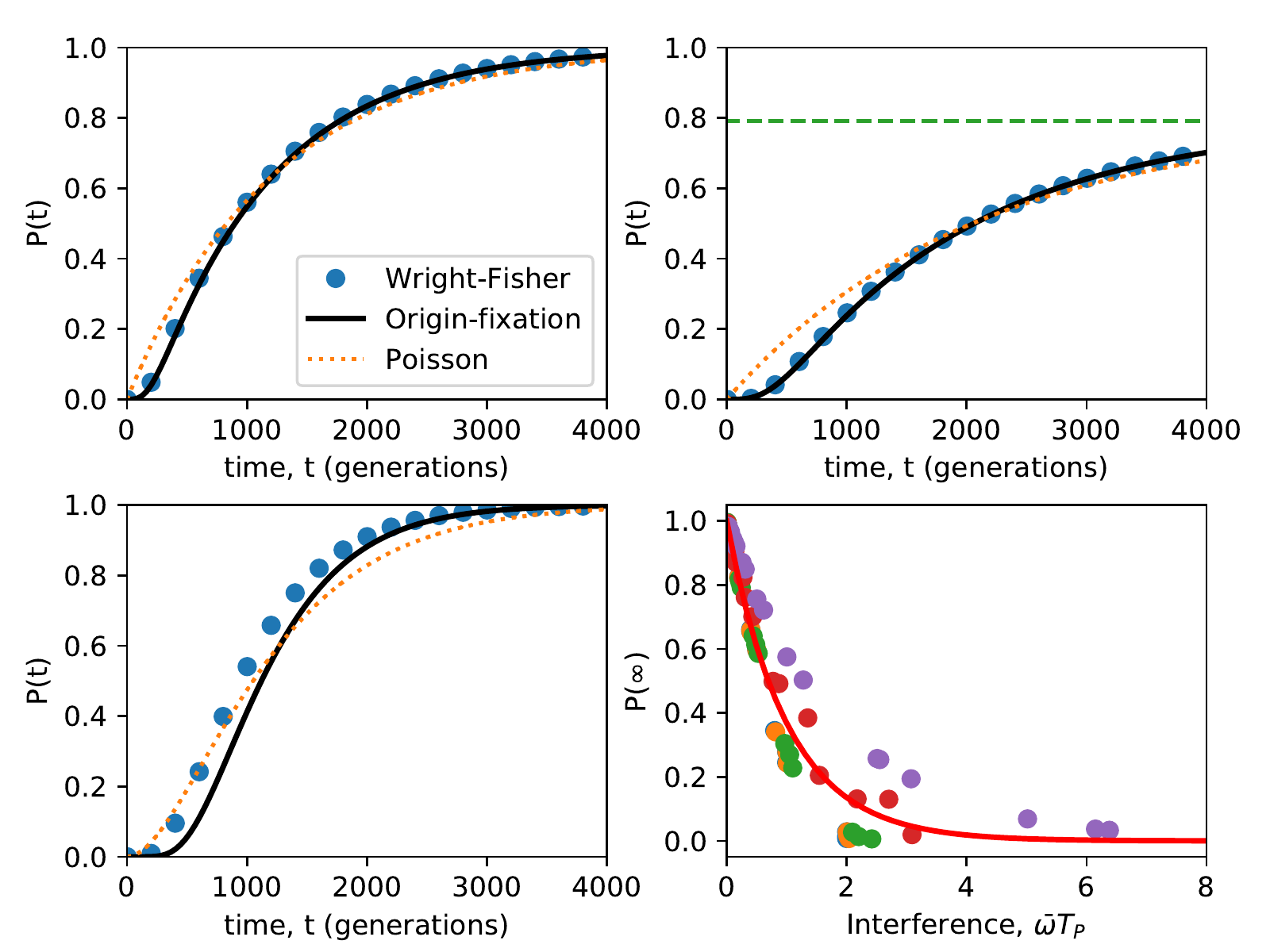}
\caption{Comparison of the origin-fixation model with numerical solutions of the full Wright-Fisher model. (Top left) Probability that a single change has occurred, $N=100$, $\eta=10^{-3}$, $s=0$. The dotted line is a Poisson process with the same mean. (Top right) Probability that the first of a sequence of two changes has occurred, $N=200$, $\eta=5\times10^{-4}$, $s=10^{-3}$. Interference between the first and second innovation means that first change does not always go to fixation. (Bottom left) Probability that the second in a sequence of two changes have occurred, $N=100$, $\eta=10^{-3}$, $s=10^{-2}$. (Bottom right) Probability that the first of two changes goes to fixation as a function of the interference $I = \omega \overline{T_F}$. This is reasonably well fit by the exponential $P(\infty)={\rm e}^{-I}$.}
\label{fig:wf}
\end{figure}

In arriving at the Kolmogorov equation (\ref{bfpe}), we made a number of approximations, in particular, that innovation (mutation) can be ignored when estimating the time to fixation. We therefore compare numerical solutions for the probability $P(t)$ that an innovation has reached fixation by time $t$ within the full Wright-Fisher dynamics against the formul\ae\ we have derived for the corresponding origin-fixation model. These results are shown in Fig.~\ref{fig:wf}, and we find the essential features are well captured. Of particular importance are  deviations from Poisson behavior that are evident at early times.  These deviations arise from the fact that an innovation takes a finite amount of time to propagate through the whole population, and are absent in classical origin-fixation models where $T_F$ is assumed to be zero. We see that even in a population of $100$ speakers, the probability that a single change has occurred is suppressed for a historically relevant time (${\sim}100$ generations, which would equate to $2,500$ years in a child-based model).  It is this property that is ultimately responsible for the very low likelihoods that are encountered in the main text.

These numerics also allow us to estimate the form of the interference correction $C_i$ introduced in Section~\ref{app:interfere}. To achieve this, we consider a generalisation to the Wright-Fisher model where two innovations can occur. If the innovation rate is fast enough, the second innovation can occur before the first has gone to fixation. A plot of the probability that the first goes to fixation for a variety of innovation rates gives us the correction factor $C_i$. We find that the numerical data are reasonably well fit by the function $C_i {\rm e}^{-\omega_i \overline{T_F}}$. As noted previously, it is not important to capture the exact form of this function, as the regime where successive innovations interfere is not empirically relevant. What is important is establishing where the correction becomes significant.

\subsection{Robustness of the Wright-Fisher model}
\label{app:demo}

It is well established in the population genetics literature (e.g.~\cite{Nordborg2019}) that the Wright-Fisher model approximates well the behaviour of a large number of evolutionary processes. In the main text, we presented one specific member from this class, since this allowed us to ascribe concrete meanings to the parameters $N_e$, $s$ and $T_M$ in terms of individual linguistic behaviour. Here, we demonstrate that many aspects of that specific model---for example, that all individuals have the same memory lifetime and that this remains constant over time, that they are equally biased in favour (or against) an innovation or do not themselves undergo processes of birth and death---are incidental. The key property is that linguistic behaviour is socially learnt, i.e., acquired from other members of the speech community, perhaps in the presence of biases.

This demonstration model comprises the following components:
\begin{itemize}
\item \textit{Agent lifetime} --- After being introduced into the population, an agent has a lifetime $d$ drawn from a Gamma distribution with mean $\mu_d$ and variance $\sigma_d^2$. It is then removed from the population after time $d$ has elapsed. If this removal causes the entire population to go extinct, a new population comprising a single individual is immediately established. (The probability of this event is very small once a steady state is reached, and is included simply to guarantee that every run of the simulation reaches the steady state).
\item \textit{Agent reproduction} --- When an agent is introduced into the population, it is also assigned a number of offspring $k\ge0$ drawn from a Geometric distribution with mean $\mu_k$. Associated with each offspring $n=1,2,\ldots,k$ is the age $b_n$ of the parent when the offspring is born. This age is drawn from a Gamma distribution with mean $\mu_b$ and variance $\sigma_b^2$. (If this age exceeds $d$, the parent's age of death, then it is discarded and the distribution resampled). When the time of birth arrives, a new agent is entered into the population with probability $1$ if the current population size $N$ is smaller than a \emph{carrying capacity} $K$, and with probability $1-\frac{K}{N\mu_k}$ otherwise. This rule prevents an unbounded exponential growth of the population, and in practice causes its size $N$ to fluctuate around $K$ in the steady state.
\item \textit{Initial interaction network} --- When an agent is born, its parent is marked as an \emph{interlocutor} (that is, someone who may influence its linguistic behaviour). The offspring also inherits each of its parent's $z$ interlocutors with probability $\frac{\mu_i}{z}$. The offspring inherits all interlocutors if $\mu_i>z$; otherwise it inherits $\mu_i$ of its parent's interlocutors on average.
\item \textit{Expansion of the interaction network} --- At the time of birth, an agent is also assigned an age $e$ drawn from a Gamma distributions with mean $\mu_e$ and variance $\sigma_e^2$. At this age, each member of the population, $n$, is assigned a weight $w_n=w_0$ if they are one of the agent's existing interlocutors, or $w_n = \exp(-h |\delta|)$ where $\delta$ is the age difference between the two agents. The existing interlocutors are then discarded, and a new set built up, with agent $n$ being marked as an interlocutor with probability $\mu_i' w_n/Z$ (capped at 1) where $Z=\sum_n w_n$. Here $\mu_i'$ is the mean number of interlocutors arising from the expansion of the interaction network. If the \emph{homophily} parameter $h=0$, then every agent in the population has the same chance of becoming an interlocutor at the time of expansion; for $h>0$, agents who are closer in age are more likely to interact after expansion than those further away in age. Thus, in this model, young agents tend to be influenced by their parents (and their parent's peer group) whereas older agents tend to influenced by their own peer group.
\item \textit{Initial linguistic behaviour} --- When an agent is created, the frequency $x$ with which it uses the innovation is inherited unchanged from its parent.
\item \textit{Rate of linguistic interactions} --- An agent participates in a linguistic interaction that modifies its behaviour at age $a$ according to a time-inhomogeneous Poisson process of intensity $R(a) {\rm d} a = R_{\infty} + (R_0 - R_{\infty}) {\rm e}^{- a/\theta}$. That is, when the agent is born, the rate of (behaviour-modifying) interactions is $R_0$, and this decays exponentially to $R_{\infty}<R_0$ with characteristic timescale $\theta$. Thus, in this model, an agent may become less liable to change their behaviour as they age. Each of the parameters $R_0$, $R_{\infty}$ and $\theta$ are assigned from distributions when the agent is born. $R_0$ is drawn from a Gamma distribution with mean $\mu_{R}$ and variance $\sigma_R^2$. $R_\infty = R_0/(1+r)$ where $r$ is drawn from a Gamma distribution with mean $\mu_r$ and variance $\sigma_r^2$. The decay time $\theta$ is drawn from a Gamma distribution with mean $\mu_\theta$ and variance $\sigma_\theta^2$. 
\item \textit{Behaviour modification in a linguistic interaction} --- When a linguistic interaction takes place, each of the agent's interlocutors is included in the interaction with probability $q$, subject to a constraint that at least one interlocutor must be present. The mean frequency of the innovation among this subset of interlocutors, $y$, is calculated. The agent then updates their innovation frequency $x$ as described in the main text: a fraction $\epsilon$ of their existing frequency is replaced with $\tau=1$ if they perceive the innovation in the interaction, and with $\tau=0$ if they perceive the convention. The probability that $\tau=1$ is $\frac{(1+\chi)y}{1+\chi y}$ where $\chi$ is a bias towards (or against) the innovation. Both the parameters $\epsilon$ and $\chi$ are randomly assigned to an agent at birth. $\epsilon$ is drawn from a Beta distribution on $[0,1]$ with mean $\mu_\epsilon$ and variance $\sigma_\epsilon^2$. The bias $\chi$ is drawn from a normal distribution with mean $\mu_\chi$ and variance $\sigma_\chi^2$.
\end{itemize}

It is evident that this model is much more complex than that described in the main text: agents are born and die, and the population size fluctuates over time; there is vertical and horizontal transmission of linguistic behaviour, the proportion of which changes during an agent's lifespan; agents can be more or less liable to changing their behaviour, and the rate at which they do so decreases over time; and they can be more less disposed to the innovation. It also has a correspondingly increased number of parameters: 23 in total; by contrast the Wright-Fisher model described in the main text has only 5 (if one excludes, as here, the possibility of an innovation being generated during the process of fixation).

To establish that the demonstration falls into the general class of Wright-Fisher models---and can be well represented with a smaller number of parameters---we use the fact that over short time increments $\Delta t$, we should find that the first two moments in the change in the frequency of the innovation over the population, $\Delta x$, are
\begin{equation}
\overline{\Delta x} = \frac{\Delta t}{T_M} s x(1-x) \quad\mbox{and}\quad 
\overline{\Delta x^2} = \frac{\Delta t}{T_M} \frac{1}{N_e}  x(1-x) \;,
\end{equation}
where $x$ is the innovation frequency, and the parameters $T_M$, $s$ and $N_e$ are as described in the main text. Here, the overlines denote averages over multiple time intervals. The crucial point is that correspondence with the Wright-Fisher model is manifested as both moments varying with frequency as $x(1-x)$.

\begin{figure}
\centering
\includegraphics[width=.9\linewidth]{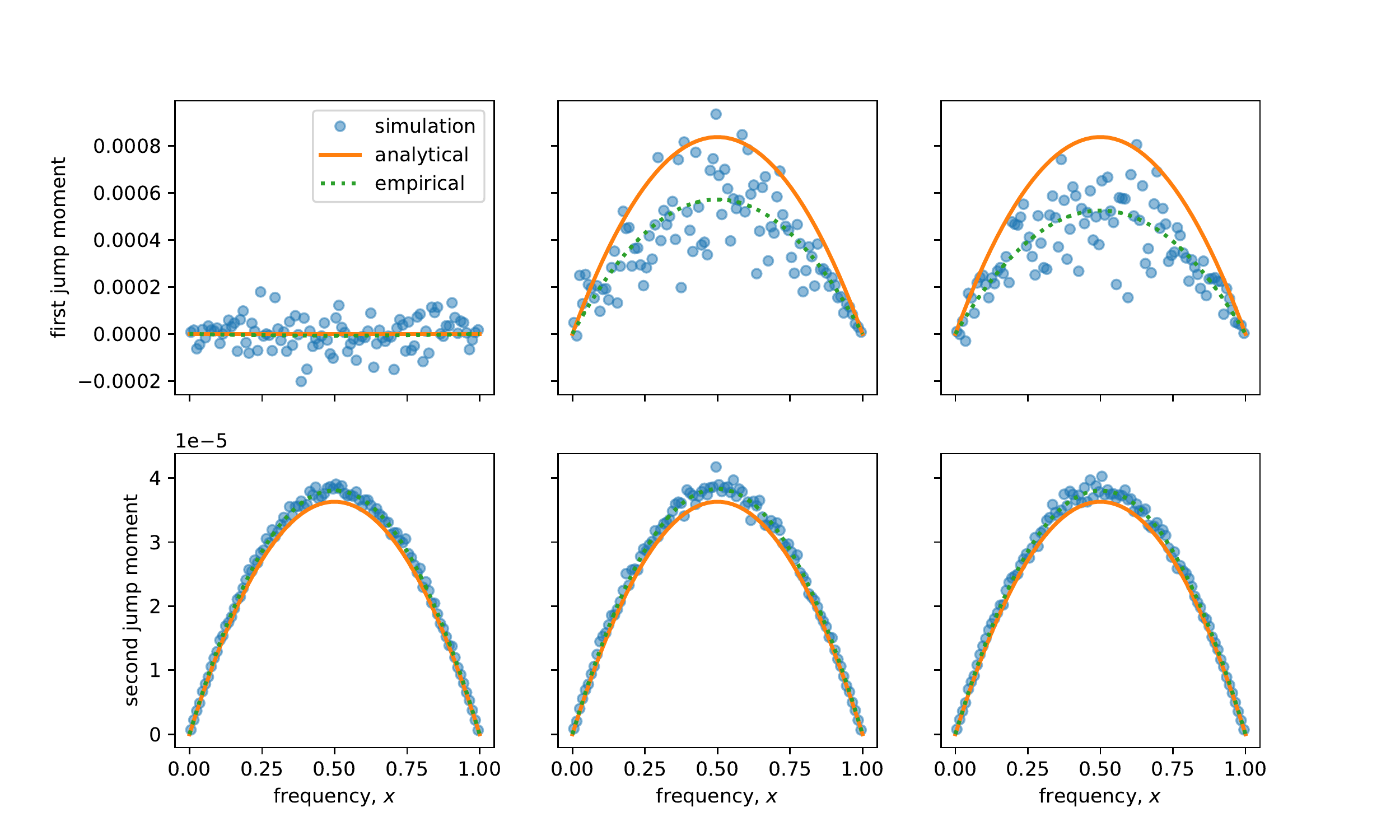}
\caption{Jump moments $\overline{\Delta x}$ and $\overline{\Delta x^2}$ obtained from simulations (points) of the demonstration model over time intervals of $\Delta t=10 {\rm yr}$. Dashed lines are empirical fits to the parabolic form $Ax(1-x)$ that corresponds to a Wright-Fisher model, where $A$ is a free parameter. Solid lines correspond to amplitudes $A$ obtained analytically through naive averaging of the relevant quantities in the demonstration model. The second jump moments are well described by these naive estimates, whilst first jump moments have the desired parabolic with an amplitude of the expected order of magnitude.}
\label{fig:jm}
\end{figure}

In Figure~\ref{fig:jm} we plot these moments obtained from simulations as a function of $x$ under three choices of the parameters controlling the bias $\chi$: (i) $\mu_\chi=0$, $\sigma_\chi=0.005$; (ii) $\mu_\chi=0.005$, $\sigma_\chi=0$; and (iii) $\mu_\chi=0.005$, $\sigma_\chi=0.005$. That is, in case (i) agents are equally likely to be biased in favour of, or against the innovation; in case (ii) all agents are biased by the same amount in favour of the innovation; and in case (iii) agents are more likely to be biased in favour of, rather than against, the innovation. The values of the remaining 21 parameters are given in Table~\ref{stab:demoparms} (and were chosen to be in the range that could plausibly describe a small human-like population). Fits of the parabola $Ax(1-x)$, with the amplitude $A$ as a free parameter, are shown as dashed lines if Figure~\ref{fig:jm}. We find these empirical fits describe well the jump moments obtained from simulation (albeit subject to some noise in the estimation of the first jump moment, which is likely a consequence of the wide variation in individual behaviour this demonstration model permits).

\begin{longtable}{l|l|l}
Parameter & Meaning & Value \\\hline
\endhead
$K$                & Carrying capacity                                                       & $1000$ \\
$\mu_k$            & Mean number of offspring                               	             & $2.0$ \\
$\mu_b$            & Mean age of parent at offspring birth                  	             & $30.0{\rm yr}$ \\
$\sigma_b$         & Standard deviation in parent age at offspring birth   		             & $8.0{\rm yr}$ \\
$\mu_e$            & Mean age at interaction network expansion                               & $18.0{\rm yr}$ \\
$\sigma_e$         & Standard deviation in age at interaction network expansion              & $4.0{\rm yr}$ \\
$\mu_d$            & Mean age at death                                                       & $60.0{\rm yr}$ \\
$\sigma_d$         & Standard deviation in age at death                                      & $12.0{\rm yr}$ \\
$\mu_i$            & Mean number of parent interlocutors inherited                           & $3.0$ \\
$\mu_i'$           & Mean number of interlocutors following expansion                        & $10.0$ \\
$h$                & Age homophily at network expansion                                      & $0.2$ \\
$w_0$	           & Inherited interlocutor weight at expansion                              & $1.0$ \\
$\mu_R$            & Mean interaction rate at birth                                          & $1.0{\rm yr}^{-1}$ \\
$\sigma_R$         & Standard deviation in interaction rate at birth                         & $0.1{\rm yr}^{-1}$ \\
$\mu_r$            & Mean interaction rate decrease with age                                 & $10.0$ \\
$\sigma_r$         & Standard deviation in interaction rate decrease with age                & $10.0$ \\
$\mu_\theta$       & Mean time over which interaction rate decays                            & $20.0{\rm yr}$ \\
$\sigma_\theta$    & Standard deviation in time over which interaction rate decays           & $10.0{\rm y}$ \\
$q$                & Probability each interlocutor participates in an interaction            & $0.5$ \\
$\mu_\epsilon$     & Mean fraction of innovation frequency that is replaced                  & $0.15$ \\
$\sigma_\epsilon$  & Standard deviation in fraction of innovation frequency that is replaced & $0.15$ \\
\caption{\label{stab:demoparms} Parameter values in the demonstration agent-based model}
\end{longtable}

Moreover, we can form naive estimates of the parameters $T_M$, $\chi$ and $N_e$ by averaging over the distributions set out above. For example, we can calculate the mean interaction rate $\overline{R}$ by averaging over the distributions of speaker lifetimes and the parameters $R_0$, $R_\infty$ and $\theta$ that govern how an individual's interaction rate changes over time. We estimate
\begin{equation}
T_M = \frac{1}{\overline{R} \overline{\epsilon}} \quad,\quad
s = \overline{\chi} \quad\mbox{and}\quad
N_e = \overline{N} \frac{\overline{\epsilon}}{\overline{\epsilon^2}} \;.
\end{equation}
The resulting parabolas are plotted as solid lines of Figure~\ref{fig:jm}. We find that the amplitude of the second jump moment (which characterises the stochastic contribution to the dynamics) is well-described by this estimate, whilst that of first jump moment is of the right order of magnitude but is over-estimated. This demonstrates that the additional complexity of this model does not fundamentally change its behaviour, but instead leads to values of the parameters in the Wright-Fisher (and therewith, the origin-fixation) model that deviate slightly from estimates obtained by simple averaging.

This observation has two important consequences for our analysis. First, by surveying all combinations of the parameters $T_M$, $s$ and $N_e$, we ultimately account for any model which---like the demonstration model here---falls into the large Wright-Fisher class. Second, in addition to the Wright-Fisher model providing a robust description of many different evolutionary processes, the interpretation of quantities like memory lifetime and individual biases given in the main text also generalises beyond the specific individual-based model presented there.


\begin{thebibliography}{100}

\bibitem{Jespersen1922}
Jespersen O.
\newblock Language, its nature, development and origin.
\newblock London: Allen and Unwin; 1922.

\bibitem{Halle1962}
Halle M.
\newblock Phonology in generative grammar.
\newblock Word. 1962;18:54--72.

\bibitem{Niyogi1997}
Niyogi P, Berwick R.
\newblock A dynamical systems model for language change.
\newblock Complex Systems. 1997;11:161--204.

\bibitem{Yang2000}
Yang CD.
\newblock Internal and external forces in language change.
\newblock Language Variation and Change. 2000;12:231--50.

\bibitem{Lightfoot2013}
Lightfoot DW.
\newblock Types of explanation in history.
\newblock Language. 2013;89:e18--e38.

\bibitem{Barlow2000}
Barlow M, Kemmer S, editors.
\newblock Usage-based models of language.
\newblock Stanford: Center for the Study of Language and Information; 2000.

\bibitem{Croft2000}
Croft W.
\newblock Explaining language change: an evolutionary approach.
\newblock Harlow, Essex: Longman; 2000.

\bibitem{Bybee2010}
Bybee JL.
\newblock Language, usage and cognition.
\newblock Cambridge: Cambridge University Press; 2010.

\bibitem{Bybee2015}
Bybee JL.
\newblock Language change.
\newblock Cambridge: Cambridge University Press; 2015.

\bibitem{Labov2001}
Labov W.
\newblock Principles of linguistic change, volume 2: Social factors.
\newblock Oxford: Wiley-Blackwell; 2001.

\bibitem{Nevalainen2003}
Nevalainen T, Raumolin-Brunberg H.
\newblock Historical sociolinguistics: language change in Tudor and Stuart
  England.
\newblock London: Routledge; 2003.

\bibitem{Sankoff2007}
Sankoff G, Blondeau H.
\newblock Language change across the lifespan: /r/ in {Montreal French}.
\newblock Language. 2007;83:566--88.

\bibitem{Baxter2016}
Baxter G, Croft W.
\newblock Modeling language change across the lifespan: Individual trajectories
  in community change.
\newblock Language Variation and Change. 2016;28:129--73.

\bibitem{Bowerman1987}
Bowerman M.
\newblock The `no negative evidence' problem: How do children avoid
  constructing an overly general grammar?
\newblock In: Hawkins JA, editor. Explaining language universals. Basil
  Blackwell; 1987. p. 73--101.

\bibitem{Tomasello2003}
Tomasello M.
\newblock Constructing a language: a usage-based theory of language
  acquisition.
\newblock Cambridge, Mass.: Harvard University Press; 2003.

\bibitem{Dressler1974}
Dressler W.
\newblock Diachronic puzzles for natural phonology.
\newblock In: Bruck A, Fox RA, LaGaly MW, editors. Papers from the Parasession
  on Natural Phonology,. Chicago: Chicago Linguistic Society; 1974. p. 95--102.

\bibitem{Drachman1978}
Drachman G.
\newblock Child language and language change: a conjectures and some
  refutations.
\newblock In: Fisiak J, editor. Recent developments in historical phonology.
  Berlin: Mouton; 1978. p. 123--44.

\bibitem{Vihman1980}
Vihman MM.
\newblock Sound change and child language.
\newblock In: Traugott EC, Labrum R, Shepherd S, editors. Papers from the 4th
  International Conference on Historical Linguistics. Amsterdam: John
  Benjamins; 1980. p. 303--20.

\bibitem{Hooper1980}
Hooper J.
\newblock Child morphology and morphophonemic change.
\newblock In: Fisiak J, editor. Historical morphology. Berlin: Mouton; 1980. p.
  157--87.

\bibitem{Bybee1982}
Bybee JL, Slobin DI.
\newblock Why small children cannot change language on their own: suggestions
  from the English past tense.
\newblock In: Ahlqvist A, editor. Papers from the 5th International Conference
  on Historical Linguistics. John Benjamins; 1982. p. 29--37.

\bibitem{Slobin1997}
Slobin DI, editor.
\newblock The crosslinguistic study of language acquisition. vol.~4.
\newblock Hillsdale, NJ: Lawrence Erlbaum Associates; 1997.

\bibitem{Ohala1989}
Ohala J.
\newblock Sound change is drawn from a pool of synchronic variation.
\newblock In: Breivik LE, Jahr EH, editors. Language change: contributions to
  the study of its causes. Berlin: Mouton de Gruyter; 1989. p. 173--98.

\bibitem{Croft2010}
Croft W.
\newblock The origins of grammaticalization in the verbalization of experience.
\newblock Linguistics. 2010;48:1--48.

\bibitem{Tagliamonte2007}
Tagliamonte SA, D'Arcy A.
\newblock Frequency and variation in the community grammar: tracking a new
  change through the generations.
\newblock Lang Var Change. 2007;19:199--217.

\bibitem{Tagliamonte2009}
Tagliamonte S, D'Arcy A.
\newblock Peaks beyond phonology: adolescence, incrementation and language
  change.
\newblock Language. 2009;85:58--108.

\bibitem{McCandlish2014}
McCandlish DM, Stoltzfus A.
\newblock Modeling evolution using the probability of fixation: History and
  implications.
\newblock The Quarterly Review of Biology. 2014;89:225--52.

\bibitem{Crow1970}
Crow JF, Kimura M.
\newblock An introduction to population genetics theory.
\newblock New York: Harper and Row; 1970.

\bibitem{WALS2013}
Dryer MS, Haspelmath M, editors.
\newblock The world atlas of language structures online.
\newblock Max Planck Institute for Evolutionary Anthropology; 2013.
\newblock Available from: \url{http://wals.info} [cited 25 March 2015].

\bibitem{Greenberg1978}
Greenberg JH.
\newblock In: Greenberg JH, Ferguson CA, Moravcsik EA, editors. How does a
  language acquire gender markers? Universals of Human Language, Vol. 3: Word
  Structure,. Stanford: Stanford University Press; 1978. p. 47--82.

\bibitem{Givon1981}
Giv\'on T.
\newblock On the development of the numeral one as an indefinite marker.
\newblock Folia Linguistica Historica. 1981;2:35--53.

\bibitem{Wichmann2009}
Wichmann S, Holman EW.
\newblock Temporal stability of linguistic typological features.
\newblock Munich: Lincom Europa; 2009.

\bibitem{Dediu2010}
Dediu D.
\newblock A {Bayesian} phylogenetic approach to estimating the stability of
  linguistic features and the genetic biasing of tone.
\newblock Proceedings of the Royal Society of London B: Biological Sciences.
  2010;doi:{10.1098/rspb.2010.1595}.

\bibitem{Kauhanen2018}
Kauhanen H, Gopal D, Galla T, Berm\'{u}dez-Otero R.
\newblock Geospatial distributions reflect rates of evolution of features of
  language.
\newblock Science Advances. 2021;7:eabe6540.

\bibitem{Pierrehumbert2003}
Pierrehumbert J.
\newblock Phonetic diversity, statistical learning, and acquisition of
  phonology.
\newblock Language and Speech. 2003;46:115--54.

\bibitem{Gillespie1983}
Gillespie JH.
\newblock Some properties of finite populations experiencing strong selection
  and weak mutation.
\newblock The American Naturalist. 1983;121:691--708.

\bibitem{Kimura1969}
Kimura M, Ohta T.
\newblock The average number of generations until fixation of a mutant gene in
  a finite population.
\newblock Genetics. 1969;61:763--71.

\bibitem{Nordborg2019}
Nordborg M.
\newblock Coalescent theory.
\newblock In: Handbook of Statistical Genomics. John Wiley; 2019. p. 145.

\bibitem{Blythe2012}
Blythe RA, Croft W.
\newblock S-curves and the mechanisms of propagation in language change.
\newblock Language. 2012;88:269--304.

\bibitem{Niyogi2009}
Niyogi P, Berwick R.
\newblock The proper treatment of language acquisition and change in a
  population setting.
\newblock PNAS. 2009;106:10124--9.

\bibitem{Lindblom1983}
Lindblom B.
\newblock Economy of speech gestures.
\newblock In: MacNeilage PF, editor. The production of speech. New York:
  Springer-Verlag; 1983. p. 217--45.

\bibitem{Christiansen2016}
Christiansen MH, Chater N.
\newblock Creating language: integrating evolution, acquisition and processing.
\newblock Cambridge, Mass.: MIT Press; 2016.

\bibitem{StClair2009}
{St Clair} MC, Monaghan P, Ramscar M.
\newblock Relationships between language structure and language learning: The
  suffixing preference and grammatical categorization.
\newblock Cognitive Science. 2009;33:1317--29.

\bibitem{Culbertson2012}
Culbertson J, Smolensky P, Legendre G.
\newblock Learning biases predict a word order universal.
\newblock Cognition. 2012;122:306--29.

\bibitem{Quine1960}
Quine WVO.
\newblock Word and object.
\newblock MIT Press; 1960.

\bibitem{Sood2005}
Sood V, Redner S.
\newblock Voter model on heterogeneous graphs.
\newblock Phy Rev Lett. 2005;94:178701.

\bibitem{Antal2006}
Antal T, Redner S, Sood V.
\newblock Evolutionary dynamics on degree-heterogeneous graphs.
\newblock Physical Review Letters. 2006;96:188104.

\bibitem{Baxter2008}
Baxter GJ, Blythe RA, McKane AJ.
\newblock Fixation and consensus times on a network: A unified approach.
\newblock Physical Review Letters. 2008;101:258701.

\bibitem{Albert2002}
Albert R, Barab{\'a}si AL.
\newblock Statistical mechanics of complex networks.
\newblock Reviews of Modern Physics. 2002;74:47--97.

\bibitem{Clauset2009}
Clauset A, Shalizi CR, Newman MEJ.
\newblock Power-law distributions in empirical data.
\newblock SIAM Review. 2009;51:661--703.

\bibitem{Kwak2010}
Kwak K, Lee C, Park H, Moon S.
\newblock What is Twitter, a social network or a news media?
\newblock In: Proceedings of the 19th international conference on {World Wide
  Web}. ACM; 2010. p. 591--600.

\bibitem{LePage1985}
LePage RB, Tabouret-Keller A.
\newblock Acts of Identity.
\newblock Cambridge: Cambridge University Press; 1985.

\bibitem{Trudgill2000}
Trudgill P.
\newblock New-dialect formation: The inevitability of colonial {Englishes}.
\newblock Edinburgh: Edinburgh University Press; 2000.

\bibitem{Baxter2009}
Baxter GJ, Blythe RA, Croft W, McKane AJ.
\newblock Modeling language change: An evaluation of {Trudgill's} theory of the
  emergence of {New Zealand English}.
\newblock Language Variation and Change. 2009;21:257--96.

\bibitem{Smith2009}
Smith K.
\newblock Iterated learning in populations of {Bayesian} agents.
\newblock In: Taatgen NA, van Rijn H, editors. Proceedings of the 31st Annual
  Conference of the Cognitive Science Society. Austin: Cognitive Science
  Society; 2009. p. 697--702.

\bibitem{Reali2010}
Reali F, Griffiths TL.
\newblock Words as alleles: connecting language evolution with {Bayesian}
  learners to models of genetic drift.
\newblock Proceedings of the Royal Society of London B: Biological Sciences.
  2010;277:429--36.

\bibitem{Burkett2010}
Burkett D, Griffiths TL.
\newblock Iterated learning of multiple languages from multiple teachers.
\newblock In: Smith ADM, Schouwstra M, de~Boer~B B, Smith K, editors. The
  Evolution of Language: Proceedings of the 8th International Conference
  (EVOLANG8). Singapore: World Scientific; 2010.

\bibitem{Baxter2006}
Baxter GJ, Blythe RA, Croft W, McKane AJ.
\newblock Utterance selection model of language change.
\newblock Physical Review E. 2006;73:046118.

\bibitem{Newberry2017}
Newberry MG, Ahern CA, Clark R, Plotkin JB.
\newblock Evolutionary forces in language change.
\newblock Nature. 2017;551:223--6.

\bibitem{Karjus2020}
Karjus A, Blythe RA, Kirby S, Smith K.
\newblock Challenges in detecting evolutionary forces in language change using
  diachronic corpora.
\newblock Glossa: a journal of general linguistics. 2020;5:45.
\newblock doi:{http://doi.org/10.5334/gjgl.909}.

\bibitem{Karsdorp2020}
Karsdorp F, Manjavacas E, Fonteyn L, Kestemont M.
\newblock Classifying evolutionary forces in language change using neural
  networks.
\newblock Evolutionary Human Sciences. 2020; p. 1--40.
\newblock doi:{10.1017/ehs.2020.52}.

\bibitem{Blythe2007}
Blythe RA, McKane AJ.
\newblock Stochastic models of evolution in genetics, ecology and linguistics.
\newblock Journal of Statistical Mechanics: Theory and Experiment. 2007; p.
  P07018.

\bibitem{code}
Blythe RA. Source code and sample datasets; 2021.
\newblock \url{https://git.ecdf.ed.ac.uk/rblythe3/gram-cycles}.

\bibitem{Burnham1998}
Burnham KP, Anderson DR.
\newblock Model selection and inference: a practical information-theoretic
  approach.
\newblock London: Springer; 1998.

\bibitem{Wichmann2008}
Wichmann S, Stauffer D, Schulze C, Holman EW.
\newblock Do language change rates depend on population size?
\newblock Advances in Complex Systems. 2008;11:357--369.

\bibitem{Lupyan2010}
Lupyan G, Dale R.
\newblock Language structure is partly determined by social structure.
\newblock PLoS ONE. 2010;5:e8559.

\bibitem{Nettle2012}
Nettle D.
\newblock Social scale and structural complexity in human languages.
\newblock Philosophical Transactions of the Royal Society of London B:
  Biological Sciences. 2012;367(1597):1829--36.

\bibitem{Bromham2015}
Bromham L, Hua X, Fitzpatrick TG, Greenhill SJ.
\newblock Rate of language evolution is affected by population size.
\newblock PNAS. 2015;112:2097--102.

\bibitem{Briscoe2000}
Briscoe R.
\newblock Grammatical acquisition: Inductive bias and coevolution of language
  and the language acquisition device.
\newblock Language. 2000;76:245--96.

\bibitem{Smith2016}
Smith K, Perfors A, Feh\'er O, Samara A, Swoboda K, Wonnacott E.
\newblock Language learning, language use and the evolution of linguistic
  variation.
\newblock Philosophical Transactions of the Royal Society of London B:
  Biological Sciences. 2016;372:20160051.
\newblock doi:{10.1098/rstb.2016.0051}.

\bibitem{Hart1995}
Hart B, Risley TR.
\newblock Meaningful differences in the everyday experience of young American
  children.
\newblock Paul H Brookes Publishing; 1995.

\bibitem{PastorSatorras2001}
Pastor-Satorras R, Vespignani A.
\newblock Epidemic spreading in scale-free networks.
\newblock Physical Review Letters. 2001;86:3200.

\bibitem{Lieberman2005}
Lieberman E, Hauert C, Nowak MA.
\newblock Evolutionary dynamics on graphs.
\newblock Nature. 2005;433:312--6.

\bibitem{Maurits2014}
Maurits L, Griffths TL.
\newblock Tracing the roots of syntax with {Bayesian} phylogenetics.
\newblock PNAS. 2014;111:13576--81.

\bibitem{Pickering2004}
Pickering MJ, Garrod S.
\newblock Toward a mechanistic psychology of dialogue.
\newblock Behavioral and Brain Sciences. 2004;27:169--226.

\bibitem{Feher2016}
Feh\'{e}r O, Wonnacott E, Smith K.
\newblock Structural priming in artificial languages and the regularisation of
  unpredictable variation.
\newblock Journal of Memory and Language. 2016;91:158--80.

\bibitem{Efferson2008}
Efferson C, Lalive R, Richerson PJ, McElreath R, Lubell M.
\newblock Conformists and mavericks: the empirics of frequency-dependent
  cultural transmission.
\newblock Evolution and Human Behaviour. 2008;29:56--64.

\bibitem{Eriksson2009}
Eriksson K, Coultas JC.
\newblock Are people really conformist-biased? An empirical test and a new
  mathematical model.
\newblock Journal of Evolutionary Psychology. 2009;7:5--21.

\bibitem{Traugott1992}
Traugott E.
\newblock Syntax.
\newblock In: Hogg RM, editor. The Cambridge History of the English Language,
  Vol. 1: The beginnings to 1066. Cambridge: Cambridge University Press; 1992.
  p. 168--289.

\bibitem{Fischer1992}
Fischer O.
\newblock Syntax.
\newblock In: Blake N, editor. The Cambridge History of the English Language,
  Vol. 2: 1066--1476. Cambridge: Cambridge University Press; 1992. p. 207--408.

\bibitem{Harbert2007}
Harbert W.
\newblock The Germanic Languages.
\newblock Cambridge: Cambridge University Press; 2007.

\bibitem{Keller1978}
Keller RE.
\newblock The German Language.
\newblock New Jersey: Humanities Press; 1978.

\bibitem{Haugen1982}
Haugen E.
\newblock Scandinavian Language Structures: A Comparative Historical Survey.
\newblock Minneapolis: University of Minnesota Press; 1982.

\bibitem{Holmes1994}
Holmes P, Hinchliffe I.
\newblock Swedish: A Comprehensive Grammar.
\newblock London: Routledge; 1994.

\bibitem{Thurneysen1946}
Thurneysen R.
\newblock A Grammar of Old Irish.
\newblock Dublin: School of Celtic Studies, Dublin Institute of Advanced
  Studies; 1946.

\bibitem{Dillon1961}
Dillon M, {{\'o} Cr{\'o}in{\'\i}n} D.
\newblock Irish.
\newblock London: English Universities Press; 1961.

\bibitem{Evans1976}
Evans DS.
\newblock A Grammar of Middle Welsh.
\newblock The Dublin Institute for Advanced Studies; 1976.

\bibitem{King2003}
King G.
\newblock Modern Welsh: A Comprehensive Grammar.
\newblock 2nd ed. London: Routledge; 2003.

\bibitem{Goodwin1892}
Goodwin WW.
\newblock A Greek Grammar.
\newblock Boston: Ginn; 1892.

\bibitem{Smyth1920}
Smyth HW.
\newblock Greek Grammar.
\newblock Cambridge, Mass.: Harvard University Press; 1920.

\bibitem{Horrocks2010}
Horrocks G.
\newblock Greek: A History of the Language and its Speakers.
\newblock Chichester: Wiley-Blackwell; 2010.

\bibitem{Holton1997}
Holton D, Mackridge P, Philippaki-Warburton I.
\newblock Greek: A Comprehensive Grammar of the Modern Language.
\newblock London: Routledge; 1997.

\bibitem{Clackson2007}
Clackson J, Horrocks G.
\newblock The Blackwell History of the Latin Language.
\newblock Chichester: Wiley-Blackwell; 2007.

\bibitem{Price1971}
Price G.
\newblock The French Language: Present and Past.
\newblock London: Edward Arnold; 1971.

\bibitem{Maiden1995}
Maiden M.
\newblock A Linguistic History of Italian.
\newblock London: Longmans; 1995.

\bibitem{Bourciez1956}
Bourciez {\'E}.
\newblock {\'E}l{\'e}ments de linguistique romane.
\newblock Paris: C. Klincksieck; 1956.

\bibitem{Huntley1993}
Huntley D.
\newblock {Old Church Slavonic}.
\newblock In: Comrie B, Corbett GG, editors. The Slavonic Languages. London:
  Routledge; 1993. p. 125--87.

\bibitem{Scatton1993}
Scatton EA.
\newblock Bulgarian.
\newblock In: Comrie B, Corbett GG, editors. The Slavonic languages. London:
  Routledge; 1993. p. 188--248.

\bibitem{Timberlake2004}
Timberlake A.
\newblock A Reference Grammar of Russian.
\newblock Cambridge: Cambridge University Press; 2004.

\bibitem{Loprieno1995}
Loprieno A.
\newblock Ancient Egyptian: A Linguistic Introduction.
\newblock Cambridge: Cambridge University Press; 1995.

\bibitem{Allen2010}
Allen JP.
\newblock The Ancient Egyptian Language: An Historical Study.
\newblock Cambridge: Cambridge University Press; 2010.

\bibitem{Junge2005}
Junge F.
\newblock Late Egyptian Grammar: An Introduction.
\newblock 2nd ed. Oxford: Griffith Institute; 2005.

\bibitem{Holes1995}
Holes C.
\newblock Modern Arabic: Structures, Functions, Varieties.
\newblock London: Longmans; 1995.

\bibitem{Lambdin1971}
Lambdin TQ.
\newblock Introduction to Biblical Hebrew.
\newblock New York: Charles Scribner's Sons; 1971.

\bibitem{Glinert1989}
Glinert L.
\newblock The Grammar of Modern Hebrew.
\newblock Cambridge: Cambridge University Press; 1989.

\bibitem{Coffin2005}
Coffin EA, Bolozky S.
\newblock A Reference Grammar of Modern Hebrew.
\newblock Cambridge: Cambridge University Press; 2005.

\bibitem{Skjaervo2005}
Skj{\ae}rv{\o} PO. An Introduction to {Old Persian}; 2005.

\bibitem{Skjaervo2009}
Skj{\ae}rv{\o} PO.
\newblock {Old Iranian}.
\newblock In: Windfuhr G, editor. The Iranian Languages. London: Routledge;
  2009. p. 43--195.

\bibitem{Lazard1992}
Lazard G.
\newblock A Grammar of Contemporary Persian.
\newblock Costa Mesa, CA: Mazda Publishers; 1992.

\bibitem{Masica1991}
Masica CF.
\newblock The Indo-Aryan Languages.
\newblock Cambridge: Cambridge University Press; 1991.

\bibitem{Sohn1999}
Sohn HM.
\newblock The Korean Language.
\newblock Cambridge: Cambridge University Press; 1999.

\bibitem{Frellesvig2010}
Frellesvig B.
\newblock A History of the Japanese Language.
\newblock Cambridge: Cambridge University Press; 2010.

\bibitem{Norman1988}
Norman J.
\newblock Chinese.
\newblock Cambridge: Cambridge University Press; 1988.

\bibitem{Launey1981}
Launey M.
\newblock Introduction {\`a} la langue et {\`a} la litt{\'e}rature
  azt{\`e}ques.
\newblock Paris: L'Harmattan; 1981.

\bibitem{Sullivan1988}
Sullivan T.
\newblock Compendium of Nahuatl Grammar.
\newblock Miller WR, Dakin K, editors. Salt Lake City: University of Utah
  Press; 1988.

\bibitem{Tuggy1979}
Tuggy DH.
\newblock {Tetelcingo Nahuatl}.
\newblock In: Langacker RW, editor. Studies in Uto-Aztecan Grammar, Vol. 2:
  Modern Aztec Grammatical Sketches. Arlington: Summer Institute of Linguistics
  and The University of Texas at Arlington; 1979. p. 1--140.

\bibitem{Brockway1979}
Brockway E.
\newblock {North Puebla Nahuatl}.
\newblock In: Langacker RW, editor. Studies in Uto-Aztecan Grammar, Vol. 2:
  Modern Aztec Grammatical Sketches. Arlington, Texas: Summer Institute of
  Linguistics and The University of Texas at Arlington; 1979. p. 141--98.

\bibitem{Beller1979}
Beller R, Beller P.
\newblock {Huasteca Nahuatl}.
\newblock In: Langacker RW, editor. Studies in Uto-Aztecan Grammar, Vol. 2:
  Modern Aztec Grammatical Sketches. Arlington, Texas: Summer Institute of
  Linguistics and The University of Texas at Arlington; 1979. p. 199--306.

\bibitem{Sischo1979}
Sischo WR.
\newblock {Michoac{\'a}n Nahuatl}.
\newblock In: Langacker RW, editor. Studies in Uto-Aztecan Grammar, Vol. 2:
  Modern Aztec Grammatical Sketches. Arlington, Texas: Summer Institute of
  Linguistics and The University of Texas at Arlington; 1979. p. 307--380.

\bibitem{McQuown1967}
McQuown NA.
\newblock {Classical Yucatec} ({Maya}).
\newblock In: McQuown NA, editor. Linguistics. vol.~5 of Handbook of Middle
  American Indians. Austin: University of Texas Press; 1967. p. 201--47.

\bibitem{Bolles1996}
Bolles D, Bolles A.
\newblock A Grammar of the Yucatecan Mayan Language.
\newblock Revised ed. Los Angeles: Foundation for the Advancement of
  Mesoamerican Studies.; 1996.

\bibitem{Edmonson1967}
Edmonson MS.
\newblock Classical {Quich{\'e}}.
\newblock In: McQuown NA, editor. Handbook of Middle American Indians, Vol. 5:
  Linguistics. Austin: University of Texas Press; 1967. p. 249--67.

\bibitem{LopezIxcoy1997}
{L{\'o}pez Ixcoy} CDS.
\newblock Ri ukemiik ri K'ichee' chi': gram{\'a}tica K'ichee'.
\newblock Guatemala, Guatemala: Cholsamaj; 1997.

\bibitem{Maxwell2006}
Maxwell JM, Hill RM.
\newblock Kaqchikel Chronicles: The Definitive Edition.
\newblock Austin: University of Texas Press; 2006.

\bibitem{Brown2006}
Brown RM, Maxwell JM, Little WE.
\newblock \textquestiondown La {\"u}tz aw{\"a}ch? Introduction to Kaqchikel
  Maya Language.
\newblock Austin: University of Texas Press; 2006.

\bibitem{Faehnrich1991}
F{\"a}hnrich H.
\newblock {Old Georgian}.
\newblock In: Harris AC, editor. The Kartvelian Languages. vol.~1 of The
  Indigenous Languages of the Caucasus. Delmar, NY: Caravan Books; 1991. p.
  129--217.

\bibitem{Tuite2004}
Tuite K.
\newblock {Early Georgian}.
\newblock In: Woodard R, editor. Encyclopedia of the World's Ancient Languages.
  Cambridge: Cambridge University Press; 2004. p. 967--87.

\bibitem{Hewitt1995}
Hewitt BG.
\newblock Georgian: A Structural Reference Grammar.
\newblock Amsterdam: John Benjamins; 1995.

\bibitem{Clackson2004}
Clackson JPT.
\newblock {Classical Armenian}.
\newblock In: Woodard R, editor. Encyclopedia of the World's Ancient Languages.
  Cambridge: Cambridge University Press; 2004. p. 922--42.

\bibitem{DumTragut2009}
Dum-Tragut J.
\newblock Armenian: Modern Eastern Armenian.
\newblock Amsterdam: John Benjamins; 2009.

\bibitem{Creason2004}
Creason S.
\newblock Aramaic.
\newblock In: Woodard R, editor. Encyclopedia of the World's Ancient Languages.
  Cambridge: Cambridge University Press; 2004. p. 381--426.

\bibitem{Lambdin1978}
Lambdin TO.
\newblock Introduction to Classical Ethiopic (Ge'ez).
\newblock Missoula, Mont.: Scholars Press; 1978.

\bibitem{Gragg1997}
Gragg G.
\newblock Ge'ez ({Ethiopic}).
\newblock In: Hetzron R, editor. The Semitic Languages. London: Routledge;
  1997. p. 242--60.

\bibitem{Faber1997}
Faber A.
\newblock Genetic subgrouping of the {Semitic} languages.
\newblock In: Hetzron R, editor. The Semitic Languages. London: Routledge;
  1997. p. 3--15.

\bibitem{Kogan1997}
Kogan LE.
\newblock Tigr{\'e}.
\newblock In: Hetzron R, editor. The Semitic Languages. London: Routledge;
  1997. p. 446--56.

\bibitem{Raz1983}
Raz S.
\newblock Tigre Grammar and Texts. vol.~4 of Afroasiatic Dialects.
\newblock Malibu, Calif: Undena Press; 1983.

\bibitem{Huehnergard2004}
Huehnergard J, Woods C.
\newblock {Akkadian} and {Eblaite}.
\newblock In: Woodard R, editor. Encyclopedia of the World's Ancient Languages.
  Cambridge: Cambridge University Press; 2004. p. 218--87.

\bibitem{Moscati1980}
Moscati S, Spitaler A, Ullendorff E, von Soden W.
\newblock An Introduction to the Comparative Grammar of the Semitic Languages:
  Phonology and Morphology.
\newblock Wiesbaden: Otto Harassowitz; 1980.

\bibitem{Michalowski2004}
Michalowski P.
\newblock Sumerian.
\newblock In: Woodard R, editor. Encyclopedia of the World's Ancient Languages.
  Cambridge: Cambridge University Press; 2004. p. 19--59.

\bibitem{Jagersma2010}
Jagersma AH.
\newblock A Descriptive Grammar of Sumerian.
\newblock Universiteit Leiden; 2010.

\bibitem{Steever2004}
Steever SB.
\newblock {Old Tamil}.
\newblock In: Woodard R, editor. Encyclopedia of the World's Ancient Languages.
  Cambridge: Cambridge University Press; 2004. p. 1015--40.

\bibitem{Rajam1992}
Rajam VS.
\newblock A Reference Grammar of Classical Tamil Poetry.
\newblock Philadelphia: American Philosophical Society; 1992.

\bibitem{Asher1982_1985}
Asher RE.
\newblock Tamil.
\newblock Croom Helm Descriptive Grammars. London: Croom Helm; 1982/1985.

\bibitem{Caldwell1875_1961}
Caldwell R.
\newblock A Comparative Grammar of the Dravidian or South-Indian Family of
  Languages.
\newblock Madras: University of Madras; 1875/1961.

\bibitem{Beyer1992}
Beyer SV.
\newblock The Classical Tibetan Language.
\newblock State University of New York Press; 1992.

\bibitem{DeLancey2003}
DeLancey S.
\newblock {Classical Tibetan}.
\newblock In: Thurgood G, LaPolla RJ, editors. The Sino-Tibetan Languages.
  London: Routledge; 2003. p. 253--69.

\bibitem{Denwood1999}
Denwood P.
\newblock Tibetan.
\newblock Amsterdam: John Benjamins; 1999.

\bibitem{Poppe1974}
Poppe N.
\newblock Grammar of Written Mongolian.
\newblock Wiesbaden: Otto Harrassowitz; 1974.

\bibitem{Binnick1979}
Binnick RI.
\newblock Modern Mongolian: A Transformational Syntax.
\newblock Toronto: University of Toronto Press; 1979.

\bibitem{Kerslake1998}
Kerslake C.
\newblock {Ottoman Turkish}.
\newblock In: Johanson L, Csat{\'o} {\'E}{\'A}, editors. The Turkic Languages.
  London: Routledge; 1998. p. 179--202.

\bibitem{Lewis1967}
Lewis GL.
\newblock Turkish Grammar.
\newblock Oxford: Oxford University Press; 1967.

\bibitem{Sidwell2009}
Sidwell P.
\newblock Classifying the Austroasiatic Languages: History and State of the
  Art.
\newblock Munich: Lincom Europa; 2009.

\bibitem{Jenner2010}
Jenner PN, Sidwell P.
\newblock Old Khmer Grammar. vol. 611 of Pacific Linguistics.
\newblock Canberra: The Australian National University; 2010.

\bibitem{Maspero1915}
Maspero G.
\newblock Grammaire de la langue khm{\`e}re (cambodgien).
\newblock Paris: Imprimerie Nationale; 1915.

\bibitem{Adelaar2004}
Adelaar WFH, Muysken PC.
\newblock The Languages of the Andes.
\newblock Cambridge: Cambridge University Press; 2004.

\bibitem{Parker1968}
Parker G.
\newblock Ayacucho Quechua Grammar and Dictionary.
\newblock The Hague: Mouton; 1968.

\bibitem{Cole1985}
Cole P.
\newblock Imbabura Quechua.
\newblock London: Croom Helm; 1985.

\bibitem{Weber1989}
Weber DJ.
\newblock A Grammar of Huallaga (Hu{\'a}naco) Quechua. vol. 112 of University
  of California Publications in Linguistics.
\newblock Berkeley: University of California Press; 1989.

\bibitem{Bertonio1603}
Bertonio PL.
\newblock Arte y grammatica muy copiosa de la lengua aymara.
\newblock Rome: Luis Zanetti; 1603.

\bibitem{Hardman2001}
Hardman MJ.
\newblock Aymara.
\newblock Munich: Lincom Europa; 2001.

\bibitem{Valdivia1060}
de~Valdivia L.
\newblock Arte y gramatica general de la lengua que corre in todo el Reyno de
  Chile, con un vocabulario, y confessionario.
\newblock Lima: Francisco del Canto; 1060.

\bibitem{Smeets2008}
Smeets I.
\newblock A Grammar of Mapuche. vol.~41 of Mouton Grammar Library.
\newblock Berlin: Mouton de Gruyter; 2008.

\bibitem{McEvedy1978}
McEvedy C, Jones R.
\newblock Atlas of World Population History.
\newblock Harmondsworth: Penguin; 1978.

\bibitem{Grimes1988}
Grimes BF.
\newblock Ethnologue.
\newblock 11th ed. Summer Institute of Linguistics; 1988.

\bibitem{Abate2006}
Abate J, Whitt W.
\newblock A Unified Framework for Numerically Inverting {Laplace} Transforms.
\newblock INFORMS Journal on Computing. 2006;18:408--21.

\end{thebibliography}
\end{document}